%% file: mamba.tex
\documentclass[sigconf]{acmart}

\AtBeginDocument{%
  }

\setcopyright{acmcopyright}
\copyrightyear{2024}
\acmYear{2024}
\acmDOI{XXXXXXX.XXXXXXX}

\acmConference[Conference acronym 'XX]{Make sure to enter the correct
  conference title from your rights confirmation emai}{June 03--05,
  2018}{Woodstock, NY}
\acmPrice{15.00}
\acmISBN{978-1-4503-XXXX-X/18/06}

\input{head.tex}

\begin{document}

\title{DTMamba : Dual Twin Mamba for Time Series Forecasting}

%

\author{Zexue Wu}
\authornote{Both authors contributed equally to this research.}
\email{zexue.wu@bit.edu.cn}
\author{Yifeng Gong}
\authornotemark[1]
\email{yifeng.gong@bit.edu.cn}
\affiliation{%
  \institution{Beijing Institute of Technology}
  \city{Beijing}
  \country{China}
}

\author{Aoqian Zhang}
\affiliation{%
  \institution{Beijing Institute of Technology}
  \country{China}}
\email{zexue.wu@bit.edu.cn}

\renewcommand{\shortauthors}{Zhang et al.}

\begin{abstract}
This is abstract.
\end{abstract}
%
%

\settopmatter{printfolios=true}
\maketitle

\input{introduction.tex}

\input{related.tex}

\input{method.tex}

\input{experiment.tex}


\section{Conclusion}
\label{sect:conclusion}

This is conclusion.


\bibliographystyle{ACM-Reference-Format}
\bibliography{mamba}

\end{document}

%% file: head.tex

\usepackage{color}
\usepackage{enumerate}
\usepackage{enumitem}
\usepackage[ruled,noend,linesnumbered]{algorithm2e}
\SetKw{Continue}{continue}
\SetKw{Break}{break}
\usepackage{url}%
\usepackage{multirow}
\usepackage[normalem]{ulem} 
\usepackage{pbox}
\usepackage{bbding}
\usepackage{makecell}
\usepackage{soul}

\newlength{\figwidths}
\newlength{\expwidths}
\newlength{\expwidthd}
\newtheorem{problem}{Problem}

\usepackage{todonotes}
\setuptodonotes{inline}



%% file: introduction.tex

\section{Introduction}
\label{sect:introduction}
%
%
%

Long-term time series forecasting (LTSF) is of paramount importance in various domains, enabling accurate predictions of future trends and patterns based on historical data. It empowers decision-makers to optimize resource allocation, make informed choices, and proactively respond to changing conditions. Accurate time series forecasting significantly benefits various domains, including finance \cite{DBLP:journals/wpc/YanO18}, supply chain management \cite{pacella2021evaluation}, healthcare \cite{DBLP:journals/tmis/MoridSD23}, and weather forecasting \cite{DBLP:journals/nn/KarevanS20}.

\par In the past, many traditional methods such as statistic-based method \cite{box1968some} have exhibited limited performance in capturing the complex and non-linear relationships between data. With the advancements in deep learning, time series forecasting has made significant progress. RNN-based \cite{DBLP:conf/ic3/MadanM18} and TCN-based \cite{DBLP:journals/soco/HewageBTPGPL20} methods, utilizing deep neural networks, have been employed to capture the underlying correlations among data, thereby enhancing the performance of predictions. However, RNN-based methods are often hindered by the issue of vanishing gradients, making it challenging to capture long-term dependencies in time series data. Additionally, RNN-based methods cannot be parallelized effectively, leading to lower computational efficiency. Furthermore, TCN-based methods primarily model the variations among adjacent time points, and they may have limited modeling capabilities for long-term dependencies \cite{DBLP:conf/iclr/WuHLZ0L23}. Especially with the advent of Transformers \cite{DBLP:conf/nips/VaswaniSPUJGKP17}, time series forecasting based on deep learning methods has reached new heights. Transformer-based methods, such as \textsf{iTransformer} \cite{DBLP:journals/corr/abs-2310-06625}, \textsf{PatchTST} \cite{DBLP:conf/iclr/NieNSK23}, and \textsf{Crossformer} \cite{DBLP:conf/iclr/ZhangY23}, have successfully leveraged the self-attention mechanisms of Transformers to capture the long-term dependencies in time series data, achieving SOTA performance. While Transformer-based methods have benefited from the advantages of Transformers, they also inherit the drawback of quadratic complexity, resulting in a significant increase in computational requirements for long input sequences. To overcome the issue of quadratic complexity in Transformer-based methods, some researchers have found that simple linear models such as \textsf{DLinear} \cite{DBLP:conf/aaai/ZengCZ023} and \textsf{TiDE} \cite{DBLP:journals/corr/abs-2304-08424} can outperform complex Transformer-based methods in terms of both performance and efficiency. However, when compared to the top-performing Transformer-based methods, these linear models still fall slightly short in terms of performance.

\par Recently, \textsf{Mamba} \cite{DBLP:journals/corr/abs-2312-00752} has emerged as an innovative linear time series modeling approach that cleverly combines the characteristics of both Recurrent Neural Networks (RNN) \cite{DBLP:journals/cogsci/Elman90} and Convolutional Neural Networks (CNN) \cite{DBLP:conf/emnlp/Kim14}, effectively addressing the computational efficiency challenges when dealing with long sequences. By leveraging the framework of State Space Models (SSM) \cite{DBLP:conf/iclr/GuGR22}, \textsf{Mamba} achieves a fusion of RNN's sequential processing capability and CNN's global information processing capability. \textsf{Mamba} introduces a selection mechanism within the SSM framework. With this selective mechanism, \textsf{Mamba} can choose to focus on essential information while filtering out irrelevant details. Consequently, \textsf{Mamba} incorporates a summary of all preceding information, enabling efficient training and inference processes. Therefore, \textsf{Mamba} successfully tackles the computational efficiency issue faced by Transformer models when handling long sequence data. Moreover, \textsf{Mamba} has demonstrated considerable potential in various domains, including graph \cite{DBLP:journals/corr/abs-2402-08678}, images \cite{DBLP:journals/corr/abs-2401-04722}, and language \cite{DBLP:journals/corr/abs-2403-19887}.

\par Given the remarkable success of \textsf{Mamba} in sequence data, it is natural to consider applying Mamba to LTSF. 
In this paper, we propose a novel model based on \textsf{Mamba} called Dual Twin Mamba (\textsf{DTMamba}). First, the input is passed through the RevIN normalization layer, which has been proven to be highly effective for tasks related to time series data \cite{DBLP:conf/iclr/KimKTPCC22}. Next is a Channel Independence layer, where \textsf{PatchTST} is the first model to utilize Channel Independence. In our experiments, we observed that Channel Independence has significant advantages over Channel Mixing. The model then consists of two TMamba blocks, proposed by us, which demonstrate exceptional capability in capturing long-term dependencies in the data. Residual networks \cite{DBLP:conf/cvpr/HeZRS16} are incorporated in the two TMamba blocks to effectively prevent gradient vanishing. Following that is a projection layer, which maps the output to the desired dimensions. Finally, a reversed Channel Independence layer and a reverse RevIN normalization layer are applied to adjust the output data format and scale to match the input data. Compared to previous SOTA models, \textsf{DTMamba} outperforms previous them in terms of performance.

\par In summary, we make the following five contributions in this paper:

\begin{itemize}
\item 
We propose \textsf{DTMamba}, which mainly consists of two TMamba blocks. By utilizing two TMamba blocks, \textsf{DTMamba} effectively captures long-term dependencies in temporal data. 
\item
We evaluate the performance of \textsf{DTMamba} and demonstrate its superiority over the previous SOTA models.
\item
We conduct ablation experiments to validate the effectiveness of channel independence and residual networks in \textsf{DTMamba}.
\item
We perform parameter sensitivity experiments to demonstrate the robustness of \textsf{DTMamba} to different parameter settings.
\item
We conducted several experiments to validate the scalability of our \textsf{DTMamba}.
\end{itemize}

%% file: related.tex

\section{Related Work}
\label{sect:related}
\paragraph{Traditional methods} Traditional methods, i.e. \cite{box1968some}, such methods generally fail to capture the long-term dependencies present in time series data, resulting in poor modeling performance.

\paragraph{RNN-based methods} RNN-based methods, i.e. \textsf{LSTNet} \cite{DBLP:conf/sigir/LaiCYL18}, struggle with vanishing gradients, hampering their capacity to capture long-term dependencies in time series data. Moreover, they have limited parallelization capabilities, resulting in decreased computational efficiency.

\paragraph{Transformer-based methods} Transformer-based methods, such as \textsf{Autoformer} \cite{DBLP:conf/nips/WuXWL21}, \textsf{FEDformer} \cite{DBLP:conf/icml/ZhouMWW0022}, \textsf{Stationary} \cite{liu2022non}, \textsf{Crossformer} \cite{DBLP:conf/iclr/ZhangY23}, \textsf{PatchTST} \cite{DBLP:conf/iclr/NieNSK23}, and \textsf{iTransformer} \cite{DBLP:journals/corr/abs-2310-06625}, leverage the self-attention mechanism of Transformers to discover dependencies between arbitrary time steps. This makes them particularly suitable for modeling long-term time series data and capturing long-term dependencies. However, these methods are often affected by the quadratic complexity of Transformers, resulting in significant time consumption.

\paragraph{Linear-based methods} Linear-based methods, i.e. \textsf{Dlinear} \cite{DBLP:conf/aaai/ZengCZ023}, \textsf{TiDE} \cite{DBLP:journals/corr/abs-2304-08424}, \textsf{Rlinear} \cite{DBLP:journals/corr/abs-2305-10721}, exhibit higher time and memory efficiency compared to Transformer-based methods. Nevertheless, when confronted with scenarios marked by high volatility and non-periodic, non-smooth patterns, Linear-based models that solely rely on past observed temporal patterns fail to deliver satisfactory performance \cite{DBLP:journals/corr/abs-2303-06053}.

\paragraph{TCN-based methods} TCN-based methods, i.e. \textsf{SCINet} \cite{DBLP:conf/nips/LiuZCXLM022}, \textsf{TimesNet} \cite{DBLP:conf/iclr/WuHLZ0L23}, have made modifications to one-dimensional convolutions by employing dilated convolutions to achieve a larger receptive field. However, they still exhibit limited modeling capabilities for long-term dependencies.

\paragraph{Applications of \textsf{Mamba}} \textsf{Mamba} is a novel architecture that stands apart from any existing frameworks, addressing the limitations of the previously mentioned methods. Currently, there are variations based on \textsf{Mamba}, such as \textsf{U-Mamba} \cite{DBLP:journals/corr/abs-2401-04722}, which has been applied to biomedical image segmentation and achieved SOTA results in related tasks. Another variant is \textsf{Vim} \cite{DBLP:journals/corr/abs-2401-09417}, which enables efficient visual representation learning, demonstrating significantly improved inference speed and memory usage compared to previous Transformer-based models. Additionally, \textsf{VideoMamba} \cite{DBLP:journals/corr/abs-2403-06977} has been developed for efficient video understanding and exhibits notable advantages in long video understanding tasks.

%% file: method.tex

\section{Proposed Method}
\label{sect:method}

\begin{figure*}[t]
  \centering
  \includegraphics[width=2\figwidths]{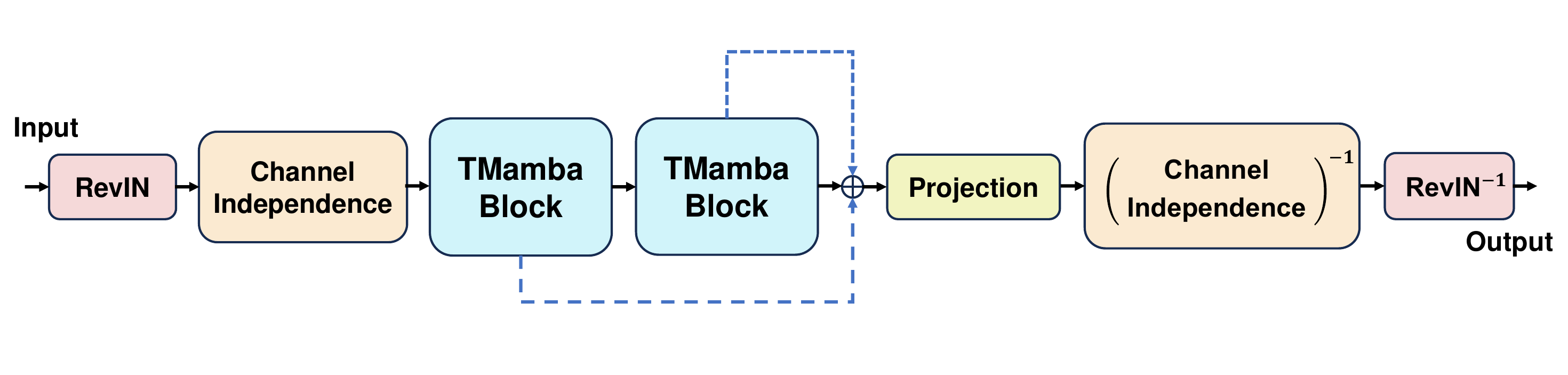}
  \caption{Overall structure of \textsf{DTMamba}.}%
  \label{fig:model}
\end{figure*}

In this section, we will provide a detailed introduction to the proposed \textsf{DTMamba}.
Figure \ref{fig:model} illustrates the overall structure of \textsf{DTMamba}, which mainly consists of three layers.
Before entering the network structure, we first normalize the time series data.
The first layer is the Channel Independence Layer, where each channel of the time series data is independently processed by our model, and channel independence is restored at the end to obtain data of the correct dimensionality.
The third layer is the \textsf{TMamba Block}, which includes Embedding, an FC layer for residual transformation, Dropout, and a pair of twin Mambas, to capture information about the variables.
The third layer is the Projection Layer, where the hidden information obtained from the preceding layers of the model is projected through FC layer to obtain the predicted values of the data.

\subsection{Problem Statement}
\label{sect:problem-statement}

\begin{definition}[Multivariate Time Series]
\label{definition:timeseries}
Consider a multivariate time series of $L$ observations, $\boldsymbol{\mathit{X}} = \{\boldsymbol{\mathit{x}}_1, \allowbreak \ldots, \boldsymbol{\mathit{x}}_L\}$. 
The $i$-th observation (data point) $\boldsymbol{\mathit{X}}_i \in \mathbb{R}^{N}$ consists of $N$ dimensions $\{\mathit{x}^1_i, \ldots, \mathit{x}^N_i\}$ and is observed at time $\mathit{t}_i$.
\end{definition}

The problem of multivariate time series data forecasting involves predicting future data of length $S$ given historical observations of length $T$.
The predicted results should ideally be as close to the true values as possible.
Generally, the predicted length $S$ of long-term time series forecasting will be greater than or equal to the historical length $T$.

\begin{problem}
Given a historical multivariate time series $\boldsymbol{\mathit{X}} = \{\boldsymbol{\mathit{x}}_1, \ldots, \boldsymbol{\mathit{x}}_T\} \in \mathbb{R}^{T \times N}$, the long-term multivariate time series forecasting problem is to find a prediction $\boldsymbol{\mathit{\hat{X}}} = \{\boldsymbol{\mathit{\hat{x}}}_{T+1}, \ldots, \boldsymbol{\mathit{\hat{x}}}_{T+S}\} \in \mathbb{R}^{S \times N}$ that is close to the true value $\boldsymbol{\mathit{X}} = \{\boldsymbol{\mathit{x}}_{T+1}, \ldots, \boldsymbol{\mathit{x}}_{T+S}\}$, where $T \leq S$.
\end{problem}

\subsection{Normalization}
\label{sect:normalization}

Before multivariate time series data $\boldsymbol{\mathit{X}}\in \mathbb{R}^{T \times N}$ enters the model, we normalize it into $\boldsymbol{\mathit{X^0}} = \{\boldsymbol{\mathit{x}}_1^0, \ldots, \boldsymbol{\mathit{x}}_T^0\} \in \mathbb{R}^{T \times N}$, via $\boldsymbol{\mathit{X^0}} = RevIN(\boldsymbol{\mathit{X}})$.
Here, $RevIN(\cdot)$ represents the normalization method, with the input being the raw data and the output being the normalized data, and $RevIN^{-1}(\cdot)$ represents the reverse normalization method.
The normalizetion operation used in this paper is the reversible instance normalization (RevIN) \cite{DBLP:conf/iclr/KimKTPCC22}.
RevIN is a normalization method aimed at improving the accuracy of time series forecasting and has also been applied in \textsf{PatchTST}.

\subsection{Channel Independence \& Reversed Channel Independence}
\label{sect:channel-independence}

Channel Independence in time series data prediction can prevent model overfitting \cite{DBLP:conf/iclr/NieNSK23}.
Therefore, before time series data enters the model, we also need to reshape it so that the data of each channel can be processed independently.
Assuming the input is $Batch(\boldsymbol{\mathit{X^0}})$, its shape is (Batch\_Size, Lookback length, Dimension number)=$(B,T,N)$.
We will reshape $Batch(\boldsymbol{\mathit{X^0}})$ into $Batch(\boldsymbol{\mathit{X^I}})$, via $Batch(\boldsymbol{\mathit{X^I}})=ChannelIndepence(Batch(\boldsymbol{\mathit{X^0}}))$.
The shape of $Batch(\boldsymbol{\mathit{X^I}})$ is $(B \times N,1,T)$.

Since we need to obtain the time series data $\boldsymbol{\mathit{\hat{X}}} \in \mathbb{R}^{S \times N}$ at the end, we need to perform the operation of reverse channel independence, which involves reshaping the data in the same manner.
Reshaping $Batch(\boldsymbol{\mathit{X^P}}):(B \times N,1,S)$ into $Batch(\boldsymbol{\mathit{\hat{X}}}):(B,S,N)$, via $Batch(\boldsymbol{\mathit{\hat{X}}})=ChannelIndepence^{-1}(Batch(\boldsymbol{\mathit{X^P}}))$, as the line 13 of Algorithm \ref{alg:mamba}.

\subsection{Twin Mamba}
\label{sect:mamba-block}

\begin{figure}[t]
  \centering
  \includegraphics[width=0.8\figwidths]{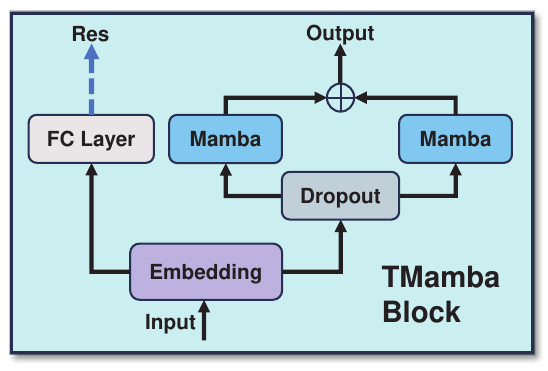}
  \caption{Details of \textsf{TMamba Block}.}%
  \label{fig:mamba-block}
\end{figure}

As shown in Figure \ref{fig:mamba-block}, upon entering the \textsf{DMamba Block}, there is first an Embedding layer, followed by a residual.
Then, it passes through a Dropout layer before entering the twin Mambas (with the same model parameter settings).
The \textsf{DMamba Block} outputs the hidden information learned by the twin Mambas and the residual.

\subsubsection{Embedding Layers}
\label{sect:embedding-layers}

Linear models exhibit unique advantages in time series prediction \cite{DBLP:journals/corr/abs-2403-14587}, so we chose to use linear layer as the Embedding layer.
Additionally, we embed the complete lookback length of the time series to obtain the global feature representation of the time series.

As the line 5 of Algorithm \ref{alg:mamba}, we will embed the $\mathit{X^I}$ into $\mathit{X^E}:(B \times N, 1, ni)$.
As shown in Figure \ref{fig:model}, \textsf{DTMamba} consists of two \textsf{TMamba Block} in total.
The Embedding layer will embed the time series data into different dimensions, namely n1 and n2.
The sensitivity of n1 and n2 will be discussed in Section \ref{sect:linear}.
Additionally, in the subsequent discussion, we will refer to the Embedding in the first \textsf{TMamba Block} as Embedding 1, and the Embedding in the second \textsf{TMamba Block} as Embedding 2.

\subsubsection{Residual}
\label{sect:residual}

The residual connection was first proposed by \textsf{ResNet} \cite{DBLP:conf/cvpr/HeZRS16}, and so far, it has been demonstrated in numerous studies to prevent overfitting, enable more stable training, and achieve better fitting \cite{DBLP:conf/eccv/HeZRS16}.
Therefore, we add a residual term after the Embedding, and to meet the dimensional requirements of the second \textsf{TMamba Block} (n2), we use a FC layer to change the dimensionality of the residual, as the line 6 of Algorithm \ref{alg:mamba}.

\subsubsection{Dropout}
\label{sect:dropout}

To prevent overfitting of the \textsf{Mamba} model to the time series data and enhance its generalization ability \cite{DBLP:journals/corr/abs-1207-0580}, Dropout is added before entering the \textsf{Mamba}.
As the line 7 of Algorithm \ref{alg:mamba}, we will change the $\mathit{X^E}$ into $\mathit{X^D}:(B \times N, 1, ni)$.

\subsubsection{Mamba}
\label{sect:mamba}

\textsf{Mamba}'s architecture is based primarily on S4, a recent state space model (SSM) architecture \cite{DBLP:journals/corr/abs-2312-00752}.
At a high level, S4 learns how to map an input $x(t)$ to an output $y(t)$ through an intermediate state $h(t)$.
Here, $x$, $y$ and $h$ are functions of $t$ because SSMs are designed to work well with continuous data such as audio, sensor data, and images.
S4 relates these to each other with three continuous parameter matrices $A$, $B$ and $C$.
These are all tied together through the following two equations:
\begin{align}
& h'(t)=Ah(t)+Bx(t) \\
& y(t)=Ch(t)
\end{align}
In practice, we always deal with discrete data, such as text.
This requires us to discretize the SSM, transforming our continuous parameters $A$, $B$ and $C$ into discrete parameters $\overline{A}$, $\overline{B}$ and $C$ by using a special fourth parameter $\Delta$:
\begin{align}
& \overline{A}=exp(\Delta A) \\
& \overline{B}=(\Delta A)^{-1}(exp(\Delta A)-I)\Delta B
\end{align}
Once discretized, we can instead represent the SSM through these two equations:
\begin{align}
& h_t=\overline{A}h_{t-1}+\overline{B}x_t \\
& y_t=Ch_t
\end{align}
These equations form a recurrence, similar to what you would see in a recurrent neural network (RNN) \cite{DBLP:journals/corr/ZarembaSV14}.
At each step $t$, we combine the hidden state from the previous timestep $h_{t-1}$ with the current input $x_t$ to create the new hidden state $h_t$.

In the \textsf{TMamba Block}, we utilize two parallel \textsf{Mamba}. Each \textsf{Mamba} model can be seen as a feature extractor, and through parallel connections, multi-level feature learning can be achieved.
One \textsf{Mamba} model can learn low-level temporal features, while the other \textsf{Mamba} model can learn high-level temporal patterns or relationships, thus enhancing the model's understanding and representation capability of time series data.

\subsection{Projection Layer}
\label{sect:projection-layer}

After obtaining the hidden information learned by the two pairs of {TMamba Block} and $\boldsymbol{\mathit{R^1}}$ and $\boldsymbol{\mathit{R^2}}$, the first step is to perform an addition operation, as indicated in line 11 of Algorithm \ref{alg:mamba}.
Then, we need to extract the time series data for the next length $S$.
We also use a linear layer to complete the projection task.
Finally, we will get $\boldsymbol{\mathit{X^P}}:(B \times N,1,S)$, which needs to undergo the reverse channel independence operation as discussed in Section \ref{sect:channel-independence} to obtain the data of the correct shape, i.e. $\boldsymbol{\mathit{\hat{X}}}:(B,S,N)$.

\begin{algorithm}[!ht]
\caption{The Forecasting Procedure of \textsf{TDMamba}}
\label{alg:mamba}
  \KwIn{$Batch(\boldsymbol{\mathit{X}})=[\boldsymbol{\mathit{x}}_1, \ldots, \boldsymbol{\mathit{x}}_T]:(B,T,N)$}
  \KwOut{$Batch(\boldsymbol{\mathit{\hat{X}}})=[\boldsymbol{\mathit{\hat{x}}}_{T+1}, \ldots, \boldsymbol{\mathit{\hat{x}}}_{T+S}]:(B,S,N)$}
	$\boldsymbol{\mathit{X^0}}:(B,T,N) \leftarrow RevIN(\boldsymbol{\mathit{X}})$ \{Normalization\}\;
	$\boldsymbol{\mathit{X^I}}:(B \times N,1,T) \leftarrow ChannelIndepence(\boldsymbol{\mathit{X^0}})$\;
	\textbf{Mamba Block}\;
	\For{$Mamba Block\_i \leftarrow Mamba Blocks$ \do}{
	$\boldsymbol{\mathit{X^E}}:(B \times N,1,ni) \leftarrow Embedding(\boldsymbol{\mathit{X^I}})$\;
	$\boldsymbol{\mathit{R^i}}:(B \times N,1,n2) \leftarrow FC(\boldsymbol{\mathit{X^E}})$ \{Residual\}\;
	$\boldsymbol{\mathit{X^D}}:(B \times N,1,ni) \leftarrow Dropout(\boldsymbol{\mathit{X^E}})$\;
	$\boldsymbol{\mathit{X^{M1}}}:(B \times N,1,ni) \leftarrow Mamba(\boldsymbol{\mathit{X^D}})$ \{low-level\}\;
	$\boldsymbol{\mathit{X^{M2}}}:(B \times N,1,ni) \leftarrow Mamba(\boldsymbol{\mathit{X^D}})$ \{high-level\}\;
	$\boldsymbol{\mathit{X^{I}}}:(B \times N,1,ni) \leftarrow \boldsymbol{\mathit{X^{M1}}} + \boldsymbol{\mathit{X^{M2}}}$
	}
	$\boldsymbol{\mathit{X^A}}:(B \times N,1,n2) \leftarrow \boldsymbol{\mathit{X^I}} + \boldsymbol{\mathit{R^1}} + \boldsymbol{\mathit{R^2}}$ \{Residual Connection\}\;
	$\boldsymbol{\mathit{X^P}}:(B \times N,1,S) \leftarrow Projection(\boldsymbol{\mathit{X^A}})$ \{Projection\}\;
	$\boldsymbol{\mathit{\hat{X}}}:(B,S,N) \leftarrow ChannelIndepence^{-1}(\boldsymbol{\mathit{X^P}})$\;
	$\boldsymbol{\mathit{\hat{X}}}:(B,S,N) \leftarrow RevIN^{-1}(\boldsymbol{\mathit{\hat{X}}})$\;
	\Return $Batch(\boldsymbol{\mathit{\hat{X}}})$\;
\end{algorithm}

%% file: experiment.tex

\section{Experiment}
\label{sect:experiment}

In this section, we conduct various time series forecasting experiments on widely recognized benchmark datasets to comprehensively evaluate the performance of our proposed model with well-acknowledged forecasting models.
The code and data of this work are available online\footnote{\url{https://anonymous.4open.science/r/mtcsc-E4CC}}.

\subsection{Settings}
All experiments were conducted using NVIDIA 2X3090 GPUs (24GB each) within the PyTorch framework \cite{DBLP:conf/nips/PaszkeGMLBCKLGA19}, with the Ubuntu 20.04 operating system.
The model was optimized using the ADAM algorithm \cite{DBLP:journals/corr/KingmaB14} with L2 regularization.
Due to the limitations of GPU memory, the batch size will be adjusted based on different datasets. However, the training for each dataset was set to 10 epochs.

\subsubsection{Datasets}
We conduct experiments on six real-world datasets widely used in the LTSF domain, including Weather, Traffic, Electricity, Exchange, four ETT (ETTh1, ETTh2, ETTm1, ETTm2) and Solar-Energy.
The detailed descriptions of the datasets are provided in Table \ref{table:dataset}.

\begin{table}[t]
 \caption{Detailed dataset descriptions.}
 \label{table:dataset}
 \centering
 \resizebox{0.8\expwidths}{!}{%
 \begin{tabular}{ccccc}
 \toprule
 Dataset & \#Dim & Time Points & Frequency & Information \\
 \midrule
 Weather \cite{DBLP:conf/nips/WuXWL21} & $21$ & $52696$ &10min& Weather \\
 Traffic \cite{DBLP:conf/nips/WuXWL21} & $862$ & $17544$ &Hourly& Transportation \\
 Electricity \cite{DBLP:conf/nips/WuXWL21} & $321$ & $26304$ &Hourly& Electricity \\
 Exchange \cite{DBLP:conf/nips/WuXWL21} & $8$ & $7588$ &Daily& Economy \\
 ETTh1 \cite{DBLP:conf/nips/WuXWL21} & $7$ & $17420$ & Hourly & Electricity \\
 ETTh2 \cite{DBLP:conf/nips/WuXWL21} & $7$ & $17420$ & Hourly & Electricity \\
 ETTm1 \cite{DBLP:conf/nips/WuXWL21} & $7$ & $69680$ & 15min & Electricity \\
 ETTm2 \cite{DBLP:conf/nips/WuXWL21} & $7$ & $69680$ & 15min & Electricity \\
 \bottomrule
 \end{tabular}
}
\end{table}

\subsubsection{Metrics}
The evaluation metrics used are Mean Squared Error (MSE) \cite{das2004mean} and Mean Absolute Error (MAE) \cite{wright1986evaluation}.
\begin{align}
& \text{MSE} = \frac{1}{n} \sum_{t=1}^{n} \lvert \boldsymbol{X}_t - \boldsymbol{\hat{X}}_t \lvert^2 \\
& \text{MAE} = \frac{1}{n} \sum_{t=1}^{n} \lvert \boldsymbol{X}_t - \boldsymbol{\hat{X}}_t \rvert
\end{align}
Here $\boldsymbol{\hat{X}}_t$ represents the predicted value at time $\mathit{t}$.
Lower values of MSE and MAE indicate more accurate prediction results.

\subsubsection{Baselines}
We compare our model \textsf{DTMamba} with 11 state-of-the-art models, including (1) Transformer-based methods: \textsf{Autoformer} \cite{DBLP:conf/nips/WuXWL21}, \textsf{FEDformer} \cite{DBLP:conf/icml/ZhouMWW0022}, \textsf{Stationary} \cite{liu2022non}, \textsf{Crossformer} \cite{DBLP:conf/iclr/ZhangY23}, \textsf{PatchTST} \cite{DBLP:conf/iclr/NieNSK23}, \textsf{iTransformer} \cite{DBLP:journals/corr/abs-2310-06625}; 
(2) Linear-based methods: \textsf{DLinear} \cite{DBLP:conf/aaai/ZengCZ023}, \textsf{TiDE} \cite{DBLP:journals/corr/abs-2304-08424}, \textsf{RLinear} \cite{DBLP:journals/corr/abs-2305-10721}; 
and (3) TCN-based methods: \textsf{SCINet} \cite{DBLP:conf/nips/LiuZCXLM022}, \textsf{TimesNet} \cite{DBLP:conf/iclr/WuHLZ0L23}.

\subsubsection{Model hyperparameters}
All experimental results of the model proposed in this paper were obtained under fixed hyperparameter settings.
Specifically, the settings were as follows: $\mathit{Dropout}=0.05$ (except Electricity, where it was set to 0.5), Embedding 1=256, Embedding 2=128, and the three parameters of Mamba: Dimension Expansion Factor=1; State Expansion Factor=256; Local Convolutional Width=2.

\subsection{Long-term Forecasting}
We conduct comprehensive time series forecasting experiments on various datasets.

\subsubsection{Comparison among our proposals}

\begin{table*}[t]
\caption{The prediction results of \textsf{DTMamba} and the benchmark models for various time series forecasting. The lookback length $\boldsymbol{T}$ is fixed at 96, and the forecast length $\boldsymbol{S \in \{96, 192, 336, 720\}}$. The best result is highlighted in bold black, and the second-best result is underlined. \textit{Avg} represent the average results across all forecast lengths.}
\label{table:result}
\resizebox{2\expwidths}{!}{
\begin{tabular}{llllllllllllllllllllllllll}
\hline
\multicolumn{2}{c|}{Models}                                              & \multicolumn{2}{c|}{\textsf{DTMamba}}                       & \multicolumn{2}{c|}{iTransformer}           & \multicolumn{2}{c|}{RLinear}       & \multicolumn{2}{c|}{PatchTST}            & \multicolumn{2}{c|}{Crossformer}            & \multicolumn{2}{c|}{TiDE}          & \multicolumn{2}{c|}{TimesNet}            & \multicolumn{2}{c|}{DLinear}                & \multicolumn{2}{c|}{SCINet}        & \multicolumn{2}{c|}{FEDformer}     & \multicolumn{2}{c|}{Stationary}    & \multicolumn{2}{c}{Autoformer}    \\ \hline
\multicolumn{2}{c|}{Metric}                                              & MSE         & \multicolumn{1}{r|}{MAE}            & MSE   & \multicolumn{1}{r|}{MAE}            & MSE   & \multicolumn{1}{r|}{MAE}   & MSE   & \multicolumn{1}{r|}{MAE}         & MSE            & \multicolumn{1}{r|}{MAE}   & MSE   & \multicolumn{1}{r|}{MAE}   & MSE         & \multicolumn{1}{r|}{MAE}   & MSE            & \multicolumn{1}{r|}{MAE}   & MSE   & \multicolumn{1}{r|}{MAE}   & MSE   & \multicolumn{1}{r|}{MAE}   & MSE   & \multicolumn{1}{r|}{MAE}   & MSE   & \multicolumn{1}{r}{MAE}   \\ \hline

\multicolumn{1}{c|}{\multirow{5}{*}{\rotatebox{90}{Weather}}} & \multicolumn{1}{c|}{96}  & {\underline{0.171}} & \multicolumn{1}{r|}{0.218} & 0.174 & \multicolumn{1}{r|}{\textbf{0.214}} & 0.192 & \multicolumn{1}{r|}{0.232} & 0.177 & \multicolumn{1}{r|}{{\underline{0.218}}} & \textbf{0.158} & \multicolumn{1}{r|}{0.23}  & 0.202 & \multicolumn{1}{r|}{0.261} & 0.172       & \multicolumn{1}{r|}{0.22}  & 0.196          & \multicolumn{1}{r|}{0.255} & 0.221 & \multicolumn{1}{r|}{0.306} & 0.217 & \multicolumn{1}{r|}{0.296} & 0.173 & \multicolumn{1}{r|}{0.223} & 0.266 & \multicolumn{1}{r}{0.336} \\
\multicolumn{1}{c|}{}                         & \multicolumn{1}{c|}{192} & 0.22        & \multicolumn{1}{r|}{{\underline{0.257}}}    & 0.221 & \multicolumn{1}{r|}{\textbf{0.254}} & 0.24  & \multicolumn{1}{r|}{0.271} & 0.225 & \multicolumn{1}{r|}{0.259}       & \textbf{0.206} & \multicolumn{1}{r|}{0.277} & 0.242 & \multicolumn{1}{r|}{0.298} & {\underline{0.219}} & \multicolumn{1}{r|}{0.261} & 0.237          & \multicolumn{1}{r|}{0.296} & 0.261 & \multicolumn{1}{r|}{0.34}  & 0.276 & \multicolumn{1}{r|}{0.336} & 0.245 & \multicolumn{1}{r|}{0.285} & 0.307 & \multicolumn{1}{r}{0.367} \\
\multicolumn{1}{c|}{}                         & \multicolumn{1}{c|}{336} & {\underline{0.274}} & \multicolumn{1}{r|}{{\textbf{0.296}}}    & 0.278 & \multicolumn{1}{r|}{\textbf{0.296}} & 0.292 & \multicolumn{1}{r|}{0.307} & 0.278 & \multicolumn{1}{r|}{\underline{0.297}}       & \textbf{0.272} & \multicolumn{1}{r|}{0.335} & 0.287 & \multicolumn{1}{r|}{0.335} & 0.28        & \multicolumn{1}{r|}{0.306} & 0.283          & \multicolumn{1}{r|}{0.335} & 0.309 & \multicolumn{1}{r|}{0.378} & 0.339 & \multicolumn{1}{r|}{0.38}  & 0.321 & \multicolumn{1}{r|}{0.338} & 0.359 & \multicolumn{1}{r}{0.395} \\
\multicolumn{1}{c|}{}                         & \multicolumn{1}{c|}{720} & {\underline{0.349}} & \multicolumn{1}{r|}{\textbf{0.346}} & 0.358 & \multicolumn{1}{r|}{0.349}          & 0.364 & \multicolumn{1}{r|}{0.353} & 0.354 & \multicolumn{1}{r|}{{\underline{0.348}}} & 0.398          & \multicolumn{1}{r|}{0.418} & 0.351 & \multicolumn{1}{r|}{0.386} & 0.365       & \multicolumn{1}{r|}{0.359} & \textbf{0.345} & \multicolumn{1}{r|}{0.381} & 0.377 & \multicolumn{1}{r|}{0.427} & 0.403 & \multicolumn{1}{r|}{0.428} & 0.414 & \multicolumn{1}{r|}{0.41}  & 0.419 & \multicolumn{1}{r}{0.428} \\ \cline{2-26} 
\multicolumn{1}{c|}{}                          & \multicolumn{1}{c|}{Avg} & \textbf{0.254} & \multicolumn{1}{r|}{{\underline{0.279}}}    & {\underline{0.258}} & \multicolumn{1}{r|}{\textbf{0.278}} & 0.272 & \multicolumn{1}{r|}{0.291} & 0.259          & \multicolumn{1}{r|}{0.281}          & 0.259 & \multicolumn{1}{r|}{0.315} & 0.271 & \multicolumn{1}{r|}{0.32}  & 0.259 & \multicolumn{1}{r|}{0.287} & 0.265          & \multicolumn{1}{r|}{0.317}       & 0.292 & \multicolumn{1}{r|}{0.363} & 0.309 & \multicolumn{1}{r|}{0.36}  & 0.288 & \multicolumn{1}{r|}{0.314} & 0.338  & 0.382                 \\ \hline

\multicolumn{1}{c|}{\multirow{5}{*}{\rotatebox{90}{Traffic}}} & \multicolumn{1}{c|}{96}  & {\underline{0.487}} & \multicolumn{1}{r|}{0.317}       & \textbf{0.395} & \multicolumn{1}{r|}{\textbf{0.268}} & 0.649 & \multicolumn{1}{r|}{0.389} & 0.544 & \multicolumn{1}{r|}{0.359} & 0.522 & \multicolumn{1}{r|}{{\underline{0.29}}}  & 0.805 & \multicolumn{1}{r|}{0.493} & 0.593 & \multicolumn{1}{r|}{0.321} & 0.65  & \multicolumn{1}{r|}{0.396} & 0.788 & \multicolumn{1}{r|}{0.499} & 0.587 & \multicolumn{1}{r|}{0.366} & 0.612 & \multicolumn{1}{r|}{0.338} & 0.613  & 0.388                 \\
\multicolumn{1}{c|}{}                         & \multicolumn{1}{c|}{192} & {\underline{0.498}} & \multicolumn{1}{r|}{{\underline{0.325}}} & \textbf{0.417} & \multicolumn{1}{r|}{\textbf{0.276}} & 0.601 & \multicolumn{1}{r|}{0.366} & 0.54  & \multicolumn{1}{r|}{0.354} & 0.53  & \multicolumn{1}{r|}{0.293}       & 0.756 & \multicolumn{1}{r|}{0.474} & 0.617 & \multicolumn{1}{r|}{0.336} & 0.598 & \multicolumn{1}{r|}{0.37}  & 0.789 & \multicolumn{1}{r|}{0.505} & 0.604 & \multicolumn{1}{r|}{0.373} & 0.613 & \multicolumn{1}{r|}{0.34}  & 0.616  & 0.382                 \\
\multicolumn{1}{c|}{}                         & \multicolumn{1}{c|}{336} & {\underline{0.511}} & \multicolumn{1}{r|}{0.334}       & \textbf{0.433} & \multicolumn{1}{r|}{\textbf{0.283}} & 0.609 & \multicolumn{1}{r|}{0.369} & 0.551 & \multicolumn{1}{r|}{0.358} & 0.558 & \multicolumn{1}{r|}{{\underline{0.305}}} & 0.762 & \multicolumn{1}{r|}{0.477} & 0.629 & \multicolumn{1}{r|}{0.336} & 0.605 & \multicolumn{1}{r|}{0.373} & 0.797 & \multicolumn{1}{r|}{0.508} & 0.621 & \multicolumn{1}{r|}{0.383} & 0.618 & \multicolumn{1}{r|}{0.328} & 0.622  & 0.337                 \\
\multicolumn{1}{c|}{}                         & \multicolumn{1}{c|}{720} & {\underline{0.533}} & \multicolumn{1}{r|}{{\underline{0.326}}} & \textbf{0.467} & \multicolumn{1}{r|}{\textbf{0.302}} & 0.647 & \multicolumn{1}{r|}{0.387} & 0.586 & \multicolumn{1}{r|}{0.375} & 0.589 & \multicolumn{1}{r|}{0.328}       & 0.719 & \multicolumn{1}{r|}{0.449} & 0.64  & \multicolumn{1}{r|}{0.35}  & 0.645 & \multicolumn{1}{r|}{0.394} & 0.841 & \multicolumn{1}{r|}{0.523} & 0.626 & \multicolumn{1}{r|}{0.382} & 0.653 & \multicolumn{1}{r|}{0.355} & 0.66   & 0.408                 \\ \cline{2-26} 
\multicolumn{1}{c|}{}                          & \multicolumn{1}{c|}{Avg} & 0.507          & \multicolumn{1}{r|}{0.326}          & \textbf{0.428} & \multicolumn{1}{r|}{\textbf{0.282}} & 0.626 & \multicolumn{1}{r|}{0.378} & {\underline{0.481}}    & \multicolumn{1}{r|}{{\underline{0.304}}}    & 0.55  & \multicolumn{1}{r|}{{\underline{0.304}}} & 0.76  & \multicolumn{1}{r|}{0.473} & 0.62  & \multicolumn{1}{r|}{0.336} & 0.625          & \multicolumn{1}{r|}{0.383}       & 0.804 & \multicolumn{1}{r|}{0.509} & 0.61  & \multicolumn{1}{r|}{0.376} & 0.624 & \multicolumn{1}{r|}{0.34}  & 0.628  & 0.379                 \\ \hline

\multicolumn{1}{c|}{\multirow{5}{*}{\rotatebox{90}{Exchange}}} & \multicolumn{1}{c|}{96}  & \textbf{0.083} & \multicolumn{1}{r|}{\textbf{0.201}} & {\underline{0.086}} & \multicolumn{1}{r|}{0.206}          & 0.093 & \multicolumn{1}{r|}{0.217} & 0.088          & \multicolumn{1}{r|}{{\underline{0.205}}}    & 0.256 & \multicolumn{1}{r|}{0.367} & 0.094 & \multicolumn{1}{r|}{0.218} & 0.107 & \multicolumn{1}{r|}{0.234} & 0.088          & \multicolumn{1}{r|}{0.218}       & 0.267 & \multicolumn{1}{r|}{0.396} & 0.148 & \multicolumn{1}{r|}{0.278} & 0.111       & \multicolumn{1}{r|}{0.237} & 0.197  & 0.323                 \\
\multicolumn{1}{c|}{}                          & \multicolumn{1}{c|}{192} & \textbf{0.173} & \multicolumn{1}{r|}{\textbf{0.295}} & 0.177       & \multicolumn{1}{r|}{{\underline{0.299}}}    & 0.184 & \multicolumn{1}{r|}{0.307} & {\underline{0.176}}    & \multicolumn{1}{r|}{{\underline{0.299}}}    & 0.47  & \multicolumn{1}{r|}{0.509} & 0.184 & \multicolumn{1}{r|}{0.307} & 0.226 & \multicolumn{1}{r|}{0.344} & {\underline{0.176}}    & \multicolumn{1}{r|}{0.315}       & 0.351 & \multicolumn{1}{r|}{0.459} & 0.271 & \multicolumn{1}{r|}{0.315} & 0.219       & \multicolumn{1}{r|}{0.335} & 0.3    & 0.369                 \\
\multicolumn{1}{c|}{}                          & \multicolumn{1}{c|}{336} & 0.346          & \multicolumn{1}{r|}{0.427}          & 0.331       & \multicolumn{1}{r|}{{\underline{0.417}}}    & 0.351 & \multicolumn{1}{r|}{0.432} & \textbf{0.301} & \multicolumn{1}{r|}{\textbf{0.397}} & 1.268 & \multicolumn{1}{r|}{0.883} & 0.349 & \multicolumn{1}{r|}{0.431} & 0.367 & \multicolumn{1}{r|}{0.448} & {\underline{0.313}}    & \multicolumn{1}{r|}{0.427}       & 1.324 & \multicolumn{1}{r|}{0.853} & 0.46  & \multicolumn{1}{r|}{0.427} & 0.421       & \multicolumn{1}{r|}{0.476} & 0.509  & 0.524                 \\
\multicolumn{1}{c|}{}                          & \multicolumn{1}{c|}{720} & 0.868          & \multicolumn{1}{r|}{0.698}          & {\underline{0.847}} & \multicolumn{1}{r|}{\textbf{0.691}} & 0.886 & \multicolumn{1}{r|}{0.714} & 0.901          & \multicolumn{1}{r|}{0.714}          & 1.767 & \multicolumn{1}{r|}{1.068} & 0.852 & \multicolumn{1}{r|}{0.698} & 0.964 & \multicolumn{1}{r|}{0.746} & \textbf{0.839} & \multicolumn{1}{r|}{{\underline{0.695}}} & 1.058 & \multicolumn{1}{r|}{0.797} & 1.195 & \multicolumn{1}{r|}{0.695} & 1.092       & \multicolumn{1}{r|}{0.769} & 1.447  & 0.941                 \\ \cline{2-26} 
\multicolumn{1}{c|}{}                          & \multicolumn{1}{c|}{Avg} & 0.368          & \multicolumn{1}{r|}{0.405}          & {\underline{0.36}}  & \multicolumn{1}{r|}{\textbf{0.403}} & 0.378 & \multicolumn{1}{r|}{0.417} & 0.367          & \multicolumn{1}{r|}{{\underline{0.404}}}    & 0.94  & \multicolumn{1}{r|}{0.707} & 0.37  & \multicolumn{1}{r|}{0.413} & 0.416 & \multicolumn{1}{r|}{0.443} & \textbf{0.354} & \multicolumn{1}{r|}{0.414}       & 0.75  & \multicolumn{1}{r|}{0.626} & 0.519 & \multicolumn{1}{r|}{0.429} & 0.461 & \multicolumn{1}{r|}{0.454} & 0.613  & 0.539                 \\ \hline

\multicolumn{1}{c|}{\multirow{5}{*}{\rotatebox{90}{Electricity}}} & \multicolumn{1}{c|}{96}  & {\underline{0.166}} & \multicolumn{1}{r|}{{\underline{0.256}}} & \textbf{0.148} & \multicolumn{1}{r|}{\textbf{0.24}}  & 0.201 & \multicolumn{1}{r|}{0.281} & 0.195 & \multicolumn{1}{r|}{0.285} & 0.219 & \multicolumn{1}{r|}{0.314} & 0.237 & \multicolumn{1}{r|}{0.329} & 0.168         & \multicolumn{1}{r|}{0.272}      & 0.197 & \multicolumn{1}{r|}{0.282} & 0.247 & \multicolumn{1}{r|}{0.345} & 0.193 & \multicolumn{1}{r|}{0.308} & 0.169       & \multicolumn{1}{r|}{0.273} & 0.201  & 0.317                 \\
\multicolumn{1}{c|}{}                             & \multicolumn{1}{c|}{192} & {\underline{0.178}} & \multicolumn{1}{r|}{{\underline{0.268}}} & \textbf{0.162} & \multicolumn{1}{r|}{\textbf{0.253}} & 0.201 & \multicolumn{1}{r|}{0.283} & 0.199 & \multicolumn{1}{r|}{0.289} & 0.231 & \multicolumn{1}{r|}{0.322} & 0.236 & \multicolumn{1}{r|}{0.33}  & 0.184         & \multicolumn{1}{r|}{0.289}      & 0.196 & \multicolumn{1}{r|}{0.285} & 0.257 & \multicolumn{1}{r|}{0.355} & 0.201 & \multicolumn{1}{r|}{0.315} & 0.182       & \multicolumn{1}{r|}{0.286} & 0.222  & 0.334                 \\
\multicolumn{1}{c|}{}                             & \multicolumn{1}{c|}{336} & {\underline{0.197}} & \multicolumn{1}{r|}{{\underline{0.289}}} & \textbf{0.178} & \multicolumn{1}{r|}{\textbf{0.269}} & 0.215 & \multicolumn{1}{r|}{0.298} & 0.215 & \multicolumn{1}{r|}{0.305} & 0.246 & \multicolumn{1}{r|}{0.337} & 0.249 & \multicolumn{1}{r|}{0.344} & 0.198         & \multicolumn{1}{r|}{0.3}        & 0.209 & \multicolumn{1}{r|}{0.301} & 0.269 & \multicolumn{1}{r|}{0.369} & 0.214 & \multicolumn{1}{r|}{0.329} & 0.2         & \multicolumn{1}{r|}{0.304} & 0.231  & 0.338                 \\
\multicolumn{1}{c|}{}                             & \multicolumn{1}{c|}{720} & 0.243       & \multicolumn{1}{r|}{0.326}       & 0.225          & \multicolumn{1}{r|}{\textbf{0.317}} & 0.257 & \multicolumn{1}{r|}{0.331} & 0.256 & \multicolumn{1}{r|}{0.337} & 0.28  & \multicolumn{1}{r|}{0.363} & 0.284 & \multicolumn{1}{r|}{0.373} & \textbf{0.22} & \multicolumn{1}{r|}{{\underline{0.32}}} & 0.245 & \multicolumn{1}{r|}{0.333} & 0.299 & \multicolumn{1}{r|}{0.39}  & 0.246 & \multicolumn{1}{r|}{0.355} & {\underline{0.222}} & \multicolumn{1}{r|}{0.321} & 0.254  & 0.361                 \\ \cline{2-26} 
\multicolumn{1}{c|}{}                             & \multicolumn{1}{c|}{Avg} & 0.196       & \multicolumn{1}{r|}{{\underline{0.285}}} & \textbf{0.178} & \multicolumn{1}{r|}{\textbf{0.27}}  & 0.219 & \multicolumn{1}{r|}{0.298} & 0.205 & \multicolumn{1}{r|}{0.29}  & 0.244 & \multicolumn{1}{r|}{0.334} & 0.251 & \multicolumn{1}{r|}{0.344} & {\underline{0.192}}   & \multicolumn{1}{r|}{0.295}      & 0.212 & \multicolumn{1}{r|}{0.3}   & 0.268 & \multicolumn{1}{r|}{0.365} & 0.214 & \multicolumn{1}{r|}{0.327} & 0.193       & \multicolumn{1}{r|}{0.296} & 0.227  & 0.338                 \\ \hline

\multicolumn{1}{c|}{\multirow{5}{*}{\rotatebox{90}{ETTh1}}} & \multicolumn{1}{c|}{96}  & 0.386 & \multicolumn{1}{r|}{\underline{0.399}} & 0.386 & \multicolumn{1}{r|}{0.405}       & 0.386          & \multicolumn{1}{r|}{{\textbf{0.395}}}    & 0.414 & \multicolumn{1}{r|}{0.419}       & 0.423 & \multicolumn{1}{r|}{0.448} & 0.479 & \multicolumn{1}{r|}{0.464} & {\underline{0.384}} & \multicolumn{1}{r|}{0.402} & 0.386 & \multicolumn{1}{r|}{0.4}   & 0.654 & \multicolumn{1}{r|}{0.599} & \textbf{0.376}          & \multicolumn{1}{r|}{0.419} & 0.513 & \multicolumn{1}{r|}{0.491} & 0.449  & 0.459                 \\
\multicolumn{1}{c|}{}                       & \multicolumn{1}{c|}{192} & {\underline{0.426}}    & \multicolumn{1}{r|}{\textbf{0.424}} & 0.441 & \multicolumn{1}{r|}{0.436}       & 0.437          & \multicolumn{1}{r|}{\textbf{0.424}} & 0.46  & \multicolumn{1}{r|}{0.445}       & 0.471 & \multicolumn{1}{r|}{0.474} & 0.525 & \multicolumn{1}{r|}{0.492} & 0.436       & \multicolumn{1}{r|}{{\underline{0.429}}} & 0.437 & \multicolumn{1}{r|}{0.432} & 0.719 & \multicolumn{1}{r|}{0.631} & \textbf{0.42}  & \multicolumn{1}{r|}{0.448} & 0.534 & \multicolumn{1}{r|}{0.504} & 0.5    & 0.482                 \\
\multicolumn{1}{c|}{}                       & \multicolumn{1}{c|}{336} & 0.48           & \multicolumn{1}{r|}{{\underline{0.45}}}     & 0.487 & \multicolumn{1}{r|}{0.458}       & {\underline{0.479}}    & \multicolumn{1}{r|}{\textbf{0.446}} & 0.501 & \multicolumn{1}{r|}{0.466}       & 0.57  & \multicolumn{1}{r|}{0.546} & 0.565 & \multicolumn{1}{r|}{0.515} & 0.491       & \multicolumn{1}{r|}{0.469}       & 0.481 & \multicolumn{1}{r|}{0.459} & 0.778 & \multicolumn{1}{r|}{0.659} & \textbf{0.459} & \multicolumn{1}{r|}{0.465} & 0.588 & \multicolumn{1}{r|}{0.535} & 0.521  & 0.496                 \\
\multicolumn{1}{c|}{}                       & \multicolumn{1}{c|}{720} & {\underline{0.484}}    & \multicolumn{1}{r|}{\textbf{0.47}}  & 0.503 & \multicolumn{1}{r|}{0.491}       & \textbf{0.481} & \multicolumn{1}{r|}{\textbf{0.47}}  & 0.5   & \multicolumn{1}{r|}{{\underline{0.488}}} & 0.653 & \multicolumn{1}{r|}{0.621} & 0.594 & \multicolumn{1}{r|}{0.558} & 0.521       & \multicolumn{1}{r|}{0.5}         & 0.519 & \multicolumn{1}{r|}{0.516} & 0.836 & \multicolumn{1}{r|}{0.699} & 0.506          & \multicolumn{1}{r|}{0.507} & 0.643 & \multicolumn{1}{r|}{0.616} & 0.514  & 0.512                 \\ \cline{2-26} 
\multicolumn{1}{c|}{}                       & \multicolumn{1}{c|}{Avg} & {\underline{0.444}}    & \multicolumn{1}{r|}{0.435} & 0.454 & \multicolumn{1}{r|}{{\underline{0.447}}} & 0.446          & \multicolumn{1}{r|}{\textbf{0.434}} & 0.469 & \multicolumn{1}{r|}{0.454}       & 0.529 & \multicolumn{1}{r|}{0.522} & 0.541 & \multicolumn{1}{r|}{0.507} & 0.458       & \multicolumn{1}{r|}{0.45}        & 0.456 & \multicolumn{1}{r|}{0.452} & 0.747 & \multicolumn{1}{r|}{0.647} & \textbf{0.44}  & \multicolumn{1}{r|}{0.46}  & 0.57  & \multicolumn{1}{r|}{0.537} & 0.496  & 0.487                 \\ \hline

\multicolumn{1}{c|}{\multirow{5}{*}{\rotatebox{90}{ETTh2}}} & \multicolumn{1}{c|}{96}  & {\underline{0.29}}     & \multicolumn{1}{r|}{{\underline{0.34}}}     & 0.297 & \multicolumn{1}{r|}{0.349} & \textbf{0.288} & \multicolumn{1}{r|}{\textbf{0.338}} & 0.302 & \multicolumn{1}{r|}{0.348} & 0.745 & \multicolumn{1}{r|}{0.584} & 0.4   & \multicolumn{1}{r|}{0.44}  & 0.34  & \multicolumn{1}{r|}{0.374} & 0.333 & \multicolumn{1}{r|}{0.387} & 0.707 & \multicolumn{1}{r|}{0.621} & 0.358 & \multicolumn{1}{r|}{0.397} & 0.476 & \multicolumn{1}{r|}{0.458} & 0.346  & 0.388                 \\
\multicolumn{1}{c|}{}                       & \multicolumn{1}{c|}{192} & \textbf{0.366} & \multicolumn{1}{r|}{{\underline{0.392}}}    & 0.38  & \multicolumn{1}{r|}{0.4}   & {\underline{0.374}}    & \multicolumn{1}{r|}{\textbf{0.39}}  & 0.388 & \multicolumn{1}{r|}{0.4}   & 0.877 & \multicolumn{1}{r|}{0.656} & 0.528 & \multicolumn{1}{r|}{0.509} & 0.402 & \multicolumn{1}{r|}{0.414} & 0.477 & \multicolumn{1}{r|}{0.476} & 0.86  & \multicolumn{1}{r|}{0.689} & 0.429 & \multicolumn{1}{r|}{0.439} & 0.512 & \multicolumn{1}{r|}{0.493} & 0.456  & 0.452                 \\
\multicolumn{1}{c|}{}                       & \multicolumn{1}{c|}{336} & \textbf{0.38}  & \multicolumn{1}{r|}{\textbf{0.409}} & 0.428 & \multicolumn{1}{r|}{0.432} & {\underline{0.415}}    & \multicolumn{1}{r|}{{\underline{0.426}}}    & 0.426 & \multicolumn{1}{r|}{0.433} & 1.043 & \multicolumn{1}{r|}{0.731} & 0.643 & \multicolumn{1}{r|}{0.571} & 0.452 & \multicolumn{1}{r|}{0.452} & 0.594 & \multicolumn{1}{r|}{0.541} & 1.000  & \multicolumn{1}{r|}{0.744} & 0.496 & \multicolumn{1}{r|}{0.487} & 0.552 & \multicolumn{1}{r|}{0.551} & 0.482  & 0.486                 \\
\multicolumn{1}{c|}{}                       & \multicolumn{1}{c|}{720} & \textbf{0.416} & \multicolumn{1}{r|}{\textbf{0.437}} & 0.427 & \multicolumn{1}{r|}{0.445} & {\underline{0.42}}     & \multicolumn{1}{r|}{{\underline{0.44}}}     & 0.431 & \multicolumn{1}{r|}{0.446} & 1.104 & \multicolumn{1}{r|}{0.763} & 0.874 & \multicolumn{1}{r|}{0.679} & 0.462 & \multicolumn{1}{r|}{0.468} & 0.831 & \multicolumn{1}{r|}{0.657} & 1.249 & \multicolumn{1}{r|}{0.838} & 0.463 & \multicolumn{1}{r|}{0.474} & 0.562 & \multicolumn{1}{r|}{0.56}  & 0.515  & 0.511                 \\ \cline{2-26} 
\multicolumn{1}{c|}{}                       & \multicolumn{1}{c|}{Avg} & \textbf{0.363} & \multicolumn{1}{r|}{\textbf{0.395}} & 0.383 & \multicolumn{1}{r|}{0.407} & {\underline{0.374}}    & \multicolumn{1}{r|}{{\underline{0.398}}}    & 0.387 & \multicolumn{1}{r|}{0.407} & 0.942 & \multicolumn{1}{r|}{0.684} & 0.611 & \multicolumn{1}{r|}{0.55}  & 0.414 & \multicolumn{1}{r|}{0.427} & 0.559 & \multicolumn{1}{r|}{0.515} & 0.954 & \multicolumn{1}{r|}{0.723} & 0.437 & \multicolumn{1}{r|}{0.449} & 0.526 & \multicolumn{1}{r|}{0.516} & 0.45   & 0.459                 \\ \hline

\multicolumn{1}{c|}{\multirow{5}{*}{\rotatebox{90}{ETTm1}}} & \multicolumn{1}{c|}{96}  & \textbf{0.325} & \multicolumn{1}{r|}{\textbf{0.36}}  & 0.334 & \multicolumn{1}{r|}{0.368} & 0.355 & \multicolumn{1}{r|}{0.376} & {\underline{0.329}}    & \multicolumn{1}{r|}{{\underline{0.367}}}    & 0.404 & \multicolumn{1}{r|}{0.426} & 0.364 & \multicolumn{1}{r|}{0.387} & 0.338       & \multicolumn{1}{r|}{0.375} & 0.345       & \multicolumn{1}{r|}{0.372} & 0.418 & \multicolumn{1}{r|}{0.438} & 0.379 & \multicolumn{1}{r|}{0.419} & 0.386 & \multicolumn{1}{r|}{0.398} & 0.505  & 0.475                 \\
\multicolumn{1}{c|}{}                       & \multicolumn{1}{c|}{192} & 0.375          & \multicolumn{1}{r|}{{\underline{0.386}}}    & 0.377 & \multicolumn{1}{r|}{0.391} & 0.391 & \multicolumn{1}{r|}{0.392} & \textbf{0.367} & \multicolumn{1}{r|}{\textbf{0.385}} & 0.45  & \multicolumn{1}{r|}{0.451} & 0.398 & \multicolumn{1}{r|}{0.404} & {\underline{0.374}} & \multicolumn{1}{r|}{0.387} & 0.38        & \multicolumn{1}{r|}{0.389} & 0.439 & \multicolumn{1}{r|}{0.45}  & 0.426 & \multicolumn{1}{r|}{0.441} & 0.459 & \multicolumn{1}{r|}{0.444} & 0.553  & 0.496                 \\
\multicolumn{1}{c|}{}                       & \multicolumn{1}{c|}{336} & \textbf{0.396} & \multicolumn{1}{r|}{\textbf{0.405}} & 0.426 & \multicolumn{1}{r|}{0.42}  & 0.424 & \multicolumn{1}{r|}{0.415} & {\underline{0.399}}    & \multicolumn{1}{r|}{{\underline{0.41}}}     & 0.532 & \multicolumn{1}{r|}{0.515} & 0.428 & \multicolumn{1}{r|}{0.425} & 0.41        & \multicolumn{1}{r|}{0.411} & 0.413       & \multicolumn{1}{r|}{0.413} & 0.49  & \multicolumn{1}{r|}{0.485} & 0.445 & \multicolumn{1}{r|}{0.459} & 0.495 & \multicolumn{1}{r|}{0.464} & 0.621  & 0.537                 \\
\multicolumn{1}{c|}{}                       & \multicolumn{1}{c|}{720} & \textbf{0.454} & \multicolumn{1}{r|}{{\underline{0.442}}}    & 0.491 & \multicolumn{1}{r|}{0.459} & 0.487 & \multicolumn{1}{r|}{0.45}  & \textbf{0.454} & \multicolumn{1}{r|}{\textbf{0.439}} & 0.666 & \multicolumn{1}{r|}{0.589} & 0.487 & \multicolumn{1}{r|}{0.461} & 0.478       & \multicolumn{1}{r|}{0.45}  & {\underline{0.474}} & \multicolumn{1}{r|}{0.453} & 0.595 & \multicolumn{1}{r|}{0.55}  & 0.543 & \multicolumn{1}{r|}{0.49}  & 0.585 & \multicolumn{1}{r|}{0.516} & 0.671  & 0.561                 \\ \cline{2-26} 
\multicolumn{1}{c|}{}                       & \multicolumn{1}{c|}{Avg} & {\underline{0.388}}    & \multicolumn{1}{r|}{\textbf{0.399}} & 0.407 & \multicolumn{1}{r|}{0.41}  & 0.414 & \multicolumn{1}{r|}{0.407} & \textbf{0.387} & \multicolumn{1}{r|}{{\underline{0.4}}}      & 0.513 & \multicolumn{1}{r|}{0.496} & 0.419 & \multicolumn{1}{r|}{0.419} & 0.4         & \multicolumn{1}{r|}{0.406} & 0.403       & \multicolumn{1}{r|}{0.407} & 0.485 & \multicolumn{1}{r|}{0.481} & 0.448 & \multicolumn{1}{r|}{0.452} & 0.481 & \multicolumn{1}{r|}{0.456} & 0.588  & 0.517                 \\ \hline

\multicolumn{1}{c|}{\multirow{5}{*}{\rotatebox{90}{ETTm2}}} & \multicolumn{1}{c|}{96}  & {\underline{0.177}}    & \multicolumn{1}{r|}{\textbf{0.259}} & 0.18  & \multicolumn{1}{r|}{{\underline{0.264}}} & 0.182       & \multicolumn{1}{r|}{0.265}          & \textbf{0.175} & \multicolumn{1}{r|}{\textbf{0.259}} & 0.287 & \multicolumn{1}{r|}{0.366} & 0.207 & \multicolumn{1}{r|}{0.305} & 0.187 & \multicolumn{1}{r|}{0.267} & 0.193 & \multicolumn{1}{r|}{0.292} & 0.286 & \multicolumn{1}{r|}{0.377} & 0.203 & \multicolumn{1}{r|}{0.287} & 0.192 & \multicolumn{1}{r|}{0.274} & 0.255  & 0.339                 \\
\multicolumn{1}{c|}{}                       & \multicolumn{1}{c|}{192} & \textbf{0.24}  & \multicolumn{1}{r|}{\textbf{0.3}}   & 0.25  & \multicolumn{1}{r|}{0.309}       & 0.246       & \multicolumn{1}{r|}{0.304}          & {\underline{0.241}}    & \multicolumn{1}{r|}{{\underline{0.302}}}    & 0.414 & \multicolumn{1}{r|}{0.492} & 0.29  & \multicolumn{1}{r|}{0.364} & 0.249 & \multicolumn{1}{r|}{0.309} & 0.284 & \multicolumn{1}{r|}{0.362} & 0.399 & \multicolumn{1}{r|}{0.445} & 0.269 & \multicolumn{1}{r|}{0.328} & 0.28  & \multicolumn{1}{r|}{0.339} & 0.281  & 0.34                  \\
\multicolumn{1}{c|}{}                       & \multicolumn{1}{c|}{336} & {\underline{0.31}}     & \multicolumn{1}{r|}{0.345}          & 0.311 & \multicolumn{1}{r|}{0.348}       & 0.307       & \multicolumn{1}{r|}{\textbf{0.342}} & \textbf{0.305} & \multicolumn{1}{r|}{{\underline{0.343}}}    & 0.597 & \multicolumn{1}{r|}{0.542} & 0.377 & \multicolumn{1}{r|}{0.422} & 0.321 & \multicolumn{1}{r|}{0.351} & 0.369 & \multicolumn{1}{r|}{0.427} & 0.637 & \multicolumn{1}{r|}{0.591} & 0.325 & \multicolumn{1}{r|}{0.366} & 0.334 & \multicolumn{1}{r|}{0.361} & 0.339  & 0.372                 \\
\multicolumn{1}{c|}{}                       & \multicolumn{1}{c|}{720} & \textbf{0.395} & \multicolumn{1}{r|}{\textbf{0.394}} & 0.412 & \multicolumn{1}{r|}{0.407}       & 0.407       & \multicolumn{1}{r|}{{\underline{0.398}}}    & {\underline{0.402}}    & \multicolumn{1}{r|}{0.4}            & 1.73  & \multicolumn{1}{r|}{1.042} & 0.558 & \multicolumn{1}{r|}{0.524} & 0.408 & \multicolumn{1}{r|}{0.403} & 0.554 & \multicolumn{1}{r|}{0.522} & 0.96  & \multicolumn{1}{r|}{0.735} & 0.421 & \multicolumn{1}{r|}{0.415} & 0.417 & \multicolumn{1}{r|}{0.413} & 0.433  & 0.432                 \\ \cline{2-26} 
\multicolumn{1}{c|}{}                       & \multicolumn{1}{c|}{Avg} & \textbf{0.281} & \multicolumn{1}{r|}{\textbf{0.325}} & 0.288 & \multicolumn{1}{r|}{0.332}       & {\underline{0.286}} & \multicolumn{1}{r|}{0.327}          & \textbf{0.281} & \multicolumn{1}{r|}{{\underline{0.326}}}    & 0.757 & \multicolumn{1}{r|}{0.61}  & 0.358 & \multicolumn{1}{r|}{0.404} & 0.291 & \multicolumn{1}{r|}{0.333} & 0.35  & \multicolumn{1}{r|}{0.401} & 0.571 & \multicolumn{1}{r|}{0.537} & 0.305 & \multicolumn{1}{r|}{0.349} & 0.306 & \multicolumn{1}{r|}{0.347} & 0.327  & 0.371                 \\ \hline

\multicolumn{2}{c|}{$1^{st}$ Count}                      & \multicolumn{1}{c}{13} & \multicolumn{1}{c|}{16} & \multicolumn{1}{c}{9} & \multicolumn{1}{c|}{16} & \multicolumn{1}{c}{2} & \multicolumn{1}{c|}{7} & \multicolumn{1}{c}{7} & \multicolumn{1}{c|}{4} & \multicolumn{1}{c}{3} & \multicolumn{1}{c|}{0} & \multicolumn{1}{c}{0} & \multicolumn{1}{c|}{0} & \multicolumn{1}{c}{1} & \multicolumn{1}{c|}{0} & \multicolumn{1}{c}{3} & \multicolumn{1}{c|}{0} & \multicolumn{1}{c}{0} & \multicolumn{1}{c|}{0} & \multicolumn{1}{c}{3} & \multicolumn{1}{c|}{0} & \multicolumn{1}{c}{0} & \multicolumn{1}{c|}{0} & \multicolumn{1}{c}{0} & \multicolumn{1}{c}{0} \\ \hline

\end{tabular}
}
\end{table*}

\begin{figure*}[t]
\centering
\begin{minipage}{0.25\textwidth}
\includegraphics[height=1.4in]{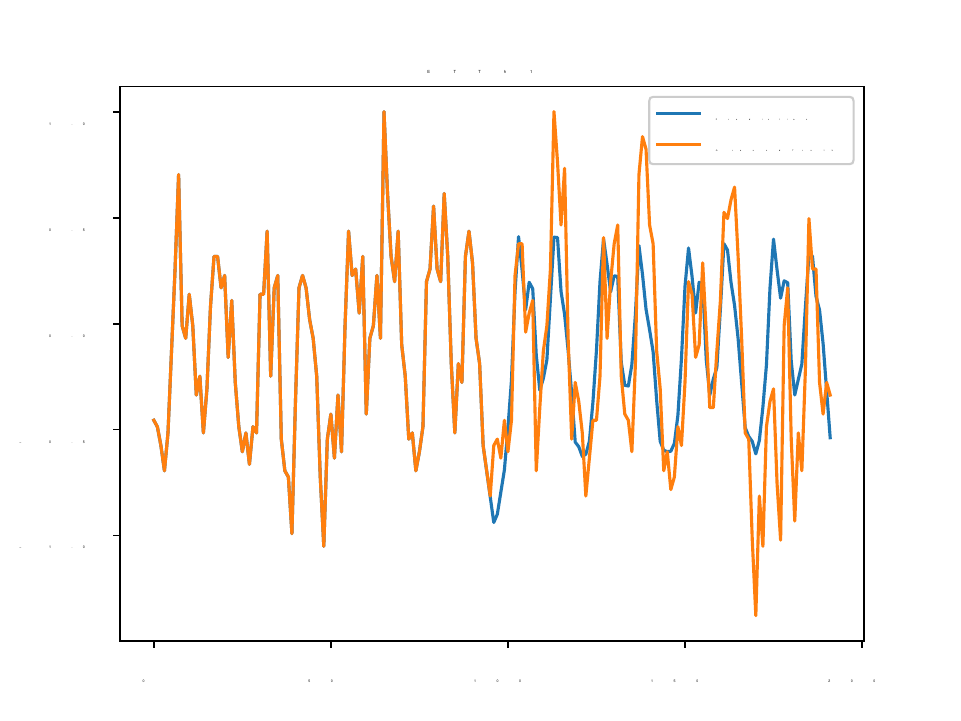}
\label{exp:mamba-ETTh1}
\end{minipage}%
\begin{minipage}{0.25\textwidth}
\centering
\includegraphics[height=1.4in]{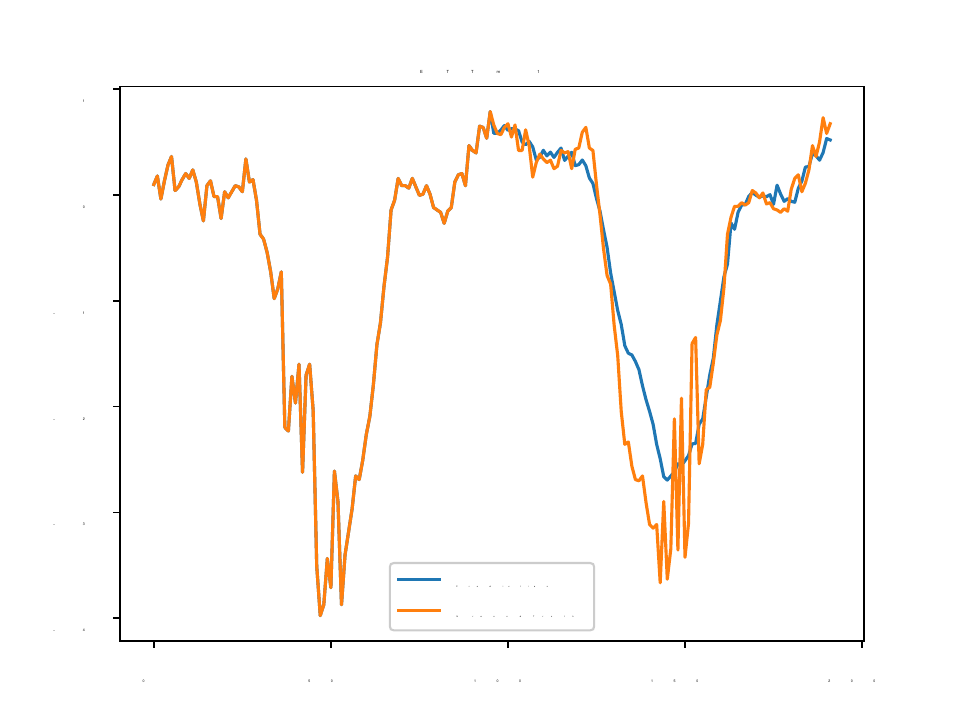}
\label{exp:mamba-ETTm1}
\end{minipage}%
\begin{minipage}{0.25\textwidth}
\centering
\includegraphics[height=1.4in]{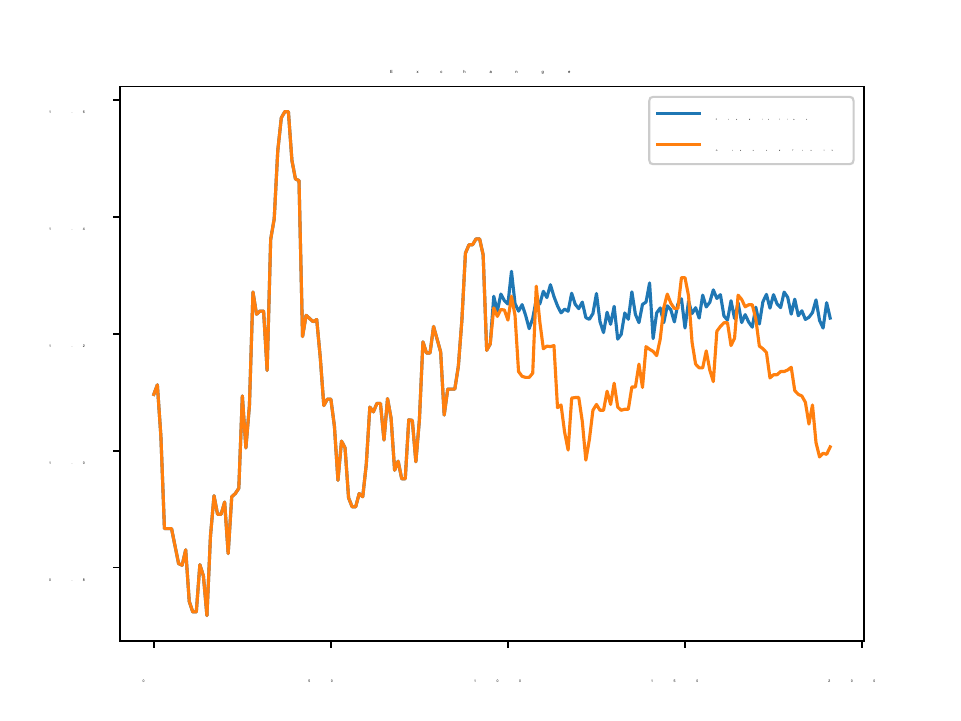}
\label{exp:mamba-exchange}
\end{minipage}%
\begin{minipage}{0.25\textwidth}
\centering
\includegraphics[height=1.4in]{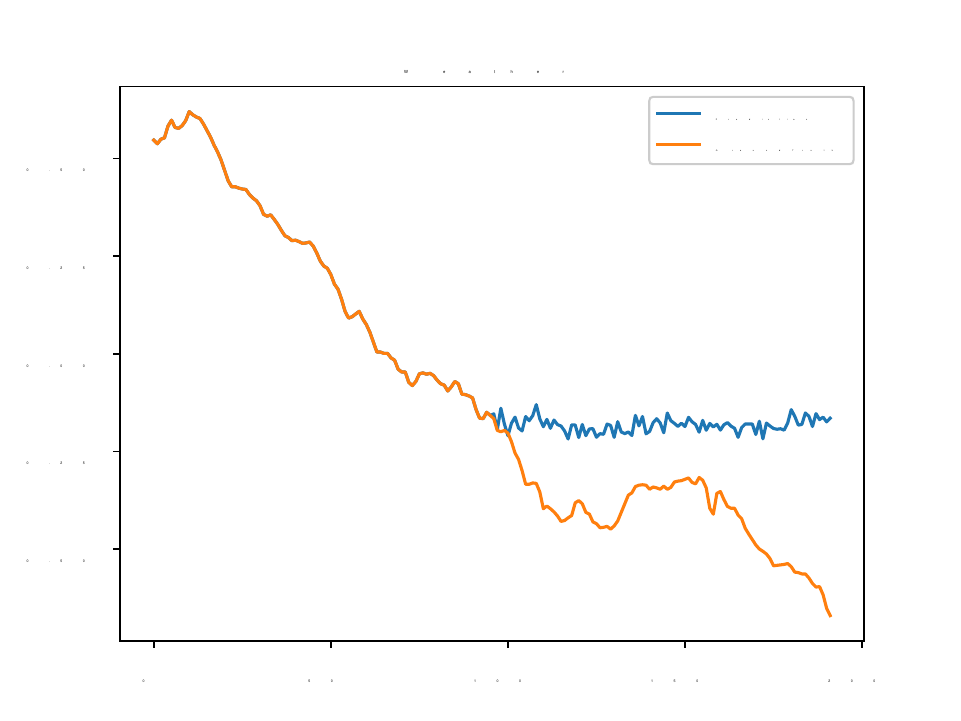}
\label{exp:mamba-weather}
\end{minipage}%

\vspace{-1em}
\caption*{(a) \textsf{DTMamba}}

\medskip

\begin{minipage}{0.25\textwidth}
\includegraphics[height=1.4in]{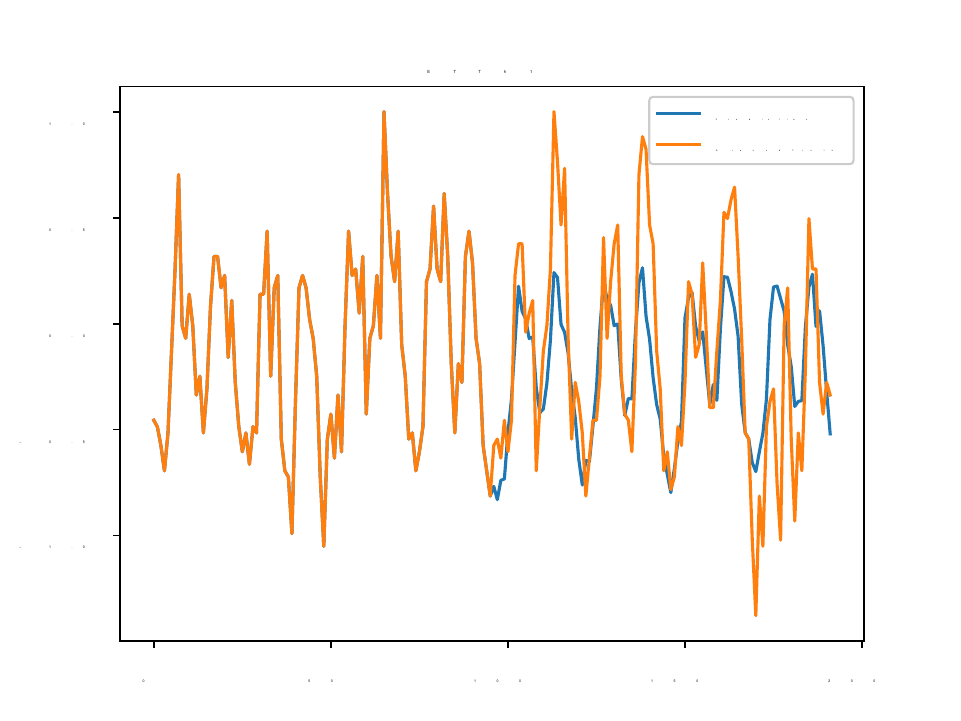}
\label{exp:iTransformer-ETTh1}
\end{minipage}%
\begin{minipage}{0.25\textwidth}
\centering
\includegraphics[height=1.4in]{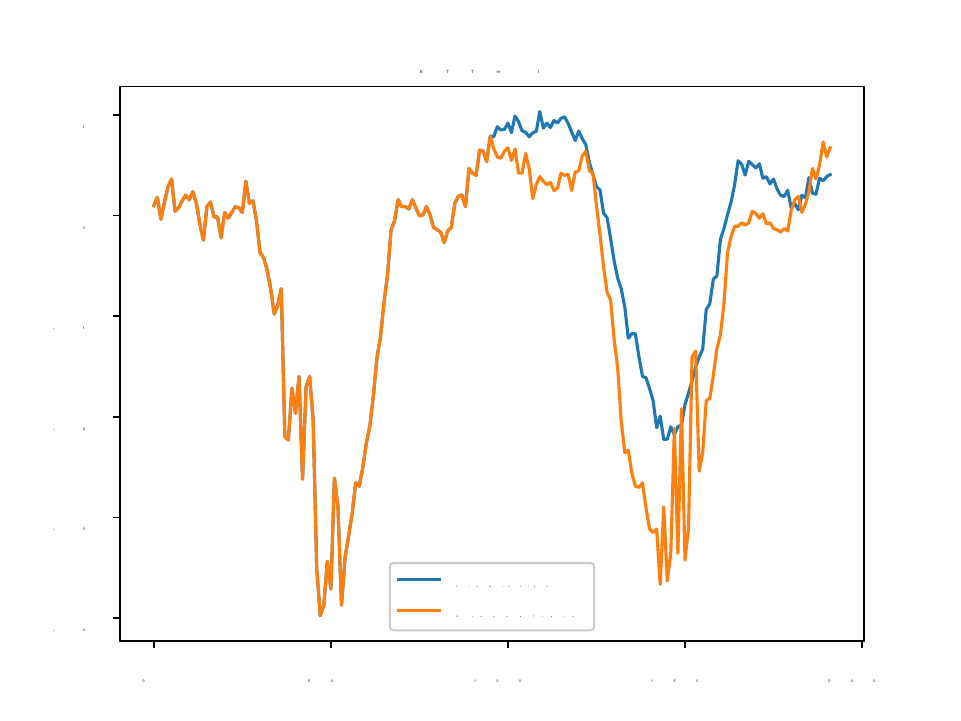}
\label{exp:iTransformer-ETTm1}
\end{minipage}%
\begin{minipage}{0.25\textwidth}
\centering
\includegraphics[height=1.4in]{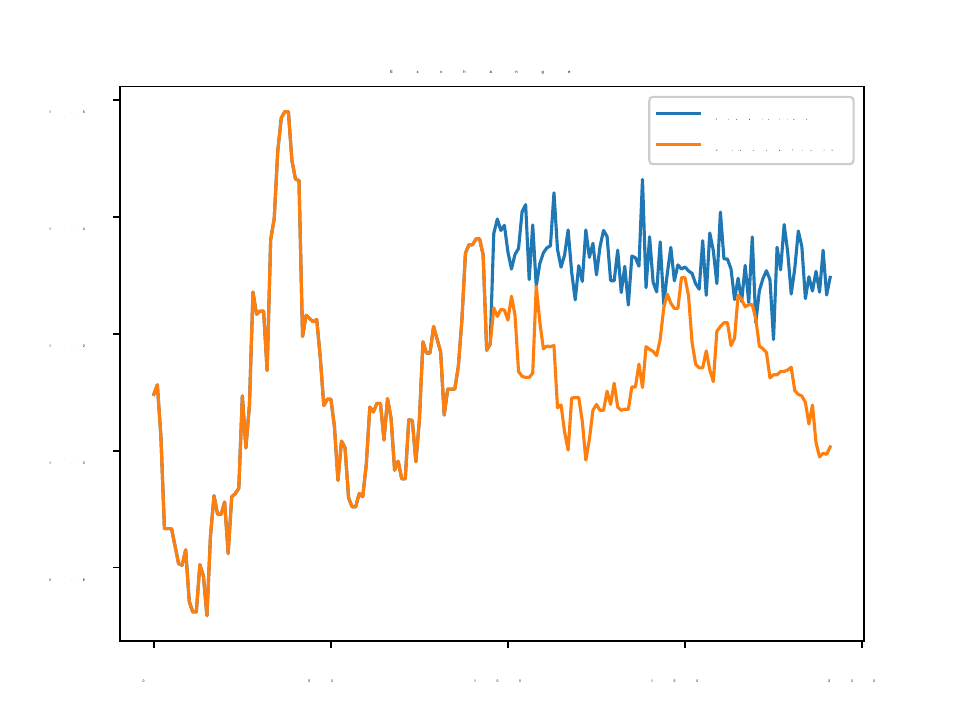}
\label{exp:iTransformer-exchange}
\end{minipage}%
\begin{minipage}{0.25\textwidth}
\centering
\includegraphics[height=1.4in]{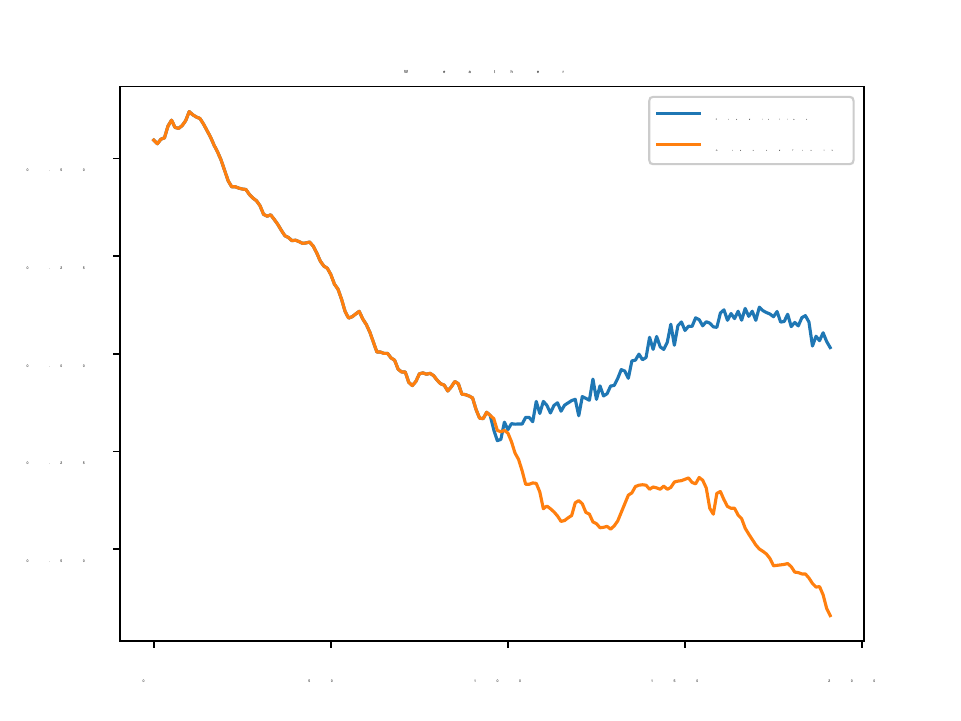}
\label{exp:iTransformer-weather}
\end{minipage}%

\vspace{-1em}
\caption*{(b) \textsf{iTransformer}}
\vspace{-1em}
\caption{The Visualization of \textsf{DTMamba} and \textsf{iTransformer} on ETTh1/ETTm1/Exchange/Weather when the lookback length $\boldsymbol{T}=96$ and the forecast length $\boldsymbol{S}=96$.}
\label{exp:visualization}
\end{figure*}


Table \ref{table:result} presents comprehensive prediction results, with the lookback length and forecast length settings consistent with mainstream LTSF.
Overall, our model demonstrates the best experimental results.
Additionally, \textsf{iTransformer} exhibits the second-best performance, showing clear advantages in the Traffic and Electricity datasets.
It is evident that our approach performs well on datasets with fewer dimensions and ranks just below \textsf{iTransformer} on datasets with higher dimensions, i.e. Traffic and Electricity.

To provide a more intuitive representation of \textsf{DTMamba}'s prediction capability, we visualize the prediction results of \textsf{DTMamba} and the second-place \textsf{iTransformer}.
We select four datasets, namely ETTh1, ETTm1, Exchange, and Weather, with a lookback length and forecast length both set to 96 (under these datasets and settings, \textsf{DTMamba}'s predictions outperform \textsf{iTransformer}).
For each dataset, we select data from one dimension.
As shown in Figure \ref{exp:visualization}, the yellow line represents the groundtruth, and the blue line represents the prediction results.
It is evident that \textsf{DTMamba}'s prediction results are closer to the groundtruth, although sometimes the performance of all models is unsatisfactory.

\subsubsection{Varying Lookback Length $\boldsymbol{T}$}

\begin{table}[t]
\caption{The prediction results of \textsf{DTMamba} with varying lookback length $\boldsymbol{T \in \{192, 336, 720\}}$ and the forecast lengths $\boldsymbol{S \in \{96, 192, 336, 720\}}$.}
\label{table:lookback}
\resizebox{\expwidths}{!}{
\begin{tabular}{cc|cc|cc|cc|cc}
\hline
\multicolumn{2}{c|}{Forecast length $\boldsymbol{S}$}           & \multicolumn{2}{c|}{96} & \multicolumn{2}{c|}{192} & \multicolumn{2}{c|}{336} & \multicolumn{2}{c}{720} \\ \hline
\multicolumn{1}{c|}{Dateset}                & T   & MSE        & MAE        & MSE         & MAE        & MSE         & MAE        & MSE        & MAE        \\ \hline
\multicolumn{1}{c|}{\multirow{3}{*}{ETTh1}} & 192 & 0.382      & 0.398      & 0.425       & 0.423      & 0.461       & 0.447      & 0.514      & 0.5        \\
\multicolumn{1}{c|}{}                       & 336 & \textbf{0.373}      & \textbf{0.397}      & \textbf{0.411}       & \textbf{0.421}      & 0.445       & \textbf{0.445}      & \textbf{0.453}      & \textbf{0.469}      \\
\multicolumn{1}{c|}{}                       & 720 & 0.386      & 0.413      & 0.414       & 0.431      & \textbf{0.441}       & 0.45       & 0.463      & 0.478      \\ \hline
\multicolumn{1}{c|}{\multirow{3}{*}{ETTh2}} & 192 & 0.297      & 0.35       & 0.369       & \textbf{0.393}      & 0.405       & 0.426      & 0.418      & \textbf{0.44}       \\
\multicolumn{1}{c|}{}                       & 336 & \textbf{0.289}      & \textbf{0.346}      & 0.375       & 0.401      & 0.379       & 0.414      & \textbf{0.412}      & 0.443      \\
\multicolumn{1}{c|}{}                       & 720 & 0.309      & 0.357      & \textbf{0.36}        & \textbf{0.393}      & \textbf{0.371}       & \textbf{0.411}      & 0.426      & 0.453      \\ \hline
\multicolumn{1}{c|}{\multirow{3}{*}{ETTm1}} & 192 & \textbf{0.296}      & 0.345      & \textbf{0.34}        & \textbf{0.374}      & \textbf{0.368}       & \textbf{0.391}      & 0.436      & 0.43       \\
\multicolumn{1}{c|}{}                       & 336 & 0.298      & \textbf{0.344}      & 0.382       & 0.405      & 0.428       & 0.431      & \textbf{0.42}       & \textbf{0.423}      \\
\multicolumn{1}{c|}{}                       & 720 & 0.346      & 0.384      & 0.35        & 0.382      & 0.423       & 0.431      & 0.433      & 0.433      \\ \hline
\multicolumn{1}{c|}{\multirow{3}{*}{ETTm2}} & 192 & 0.177      & \textbf{0.258}      & \textbf{0.233}       & \textbf{0.295}      & 0.3         & 0.342      & 0.386      & 0.393      \\
\multicolumn{1}{c|}{}                       & 336 & \textbf{0.174}      & 0.263      & 0.235       & 0.302      & 0.294       & \textbf{0.337}      & \textbf{0.363}      & \textbf{0.385}      \\
\multicolumn{1}{c|}{}                       & 720 & 0.176      & 0.264      & 0.24        & 0.314      & \textbf{0.293}       & 0.348      & 0.371      & 0.397      \\ \hline
\end{tabular}
}
\end{table}

In addition to the mainstream lookback length of 96, we also conduct experiments on our \textsf{DTMamba} with lookback lengths of 192, 336, and 720.
As shown in Table \ref{table:lookback}, MSE and MAE do not necessarily decrease with the increase in the lookback length at various forecasting lengths.
Similar observations were also made in Transformer-based models \cite{DBLP:conf/iclr/NieNSK23} \cite{DBLP:conf/aaai/ZengCZ023}.

\subsection{Hyperparameter Sensitivity Analysis and Ablation Study}
\label{sect:hyperparameter}

In this subsection, we will conduct ablation experiments on certain modules within the network structure and test the sensitivity of the network structure to various parameters in Dropout, linear layers, and \textsf{Mamba}.

\subsubsection{Ablation of Residual Connections}

\begin{table*}[t]
\caption{Residual connection and Channel Independence ablation experiment with an lookback length $\boldsymbol{T} = 96$}
\label{table:residual-channel}
\resizebox{1.5\expwidths}{!}{
\begin{tabular}{c|c|c|cc|cc|cc|cc}
\hline
\multirow{2}{*}{Residual} & \multirow{2}{*}{\begin{tabular}[c]{@{}c@{}}Channel\\ Indepence\end{tabular}} & \multirow{2}{*}{\begin{tabular}[c]{@{}c@{}}Forecast\\ length $S$\end{tabular}} & \multicolumn{2}{c|}{ETTh1}      & \multicolumn{2}{c|}{ETTh2}      & \multicolumn{2}{c|}{ETTm1}      & \multicolumn{2}{c}{ETTm2}       \\ \cline{4-11} 
                          &                                                                              &                                                                              & MSE            & MAE            & MSE            & MAE            & MSE            & MAE            & MSE            & MAE            \\ \hline
\multirow{5}{*}{Residual} & \multirow{5}{*}{\begin{tabular}[c]{@{}c@{}}Channel\\ Indepence\end{tabular}} & 96                                                                           & 0.386          & 0.399          & 0.290          & 0.340          & 0.325          & 0.360          & 0.177          & 0.259          \\
                          &                                                                              & 192                                                                          & 0.426          & 0.424          & 0.366          & 0.392          & 0.375          & 0.386          & 0.240          & 0.300          \\
                          &                                                                              & 336                                                                          & 0.480          & 0.450          & 0.380          & 0.409          & 0.396          & 0.405          & 0.310          & 0.345          \\
                          &                                                                              & 720                                                                          & 0.484          & 0.470          & 0.416          & 0.437          & 0.454          & 0.442          & 0.395          & 0.394          \\ \cline{3-11} 
                          &                                                                              & Avg                                                                          & \textbf{0.444} & \textbf{0.435} & \textbf{0.363} & \textbf{0.395} & \textbf{0.388} & \textbf{0.399} & \textbf{0.281} & \textbf{0.325} \\ \hline
\multirow{5}{*}{w/o}      & \multirow{5}{*}{\begin{tabular}[c]{@{}c@{}}Channel\\ Indepence\end{tabular}} & 96                                                                           & 0.384          & 0.408          & 0.294          & 0.345          & 0.329          & 0.367          & 0.181          & 0.263          \\
                          &                                                                              & 192                                                                          & 0.438          & 0.437          & 0.373          & 0.396          & 0.379          & 0.393          & 0.243          & 0.304          \\
                          &                                                                              & 336                                                                          & 0.482          & 0.455          & 0.396          & 0.418          & 0.402          & 0.411          & 0.307          & 0.345          \\
                          &                                                                              & 720                                                                          & 0.506          & 0.485          & 0.420          & 0.439          & 0.467          & 0.451          & 0.394          & 0.395          \\ \cline{3-11} 
                          &                                                                              & Avg                                                                          & 0.453          & 0.446          & 0.371          & 0.400          & 0.394          & 0.406          & \textbf{0.281} & 0.327          \\ \hline
\multirow{5}{*}{Residual} & \multirow{5}{*}{w/o}                                                         & 96                                                                           & 0.396          & 0.408          & 0.290          & 0.343          & 0.376          & 0.395          & 0.180          & 0.262          \\
                          &                                                                              & 192                                                                          & 0.428          & 0.426          & 0.387          & 0.400          & 0.373          & 0.413          & 0.241          & 0.304          \\
                          &                                                                              & 336                                                                          & 0.487          & 0.457          & 0.382          & 0.409          & 0.416          & 0.411          & 0.302          & 0.342          \\
                          &                                                                              & 720                                                                          & 0.495          & 0.482          & 0.426          & 0.441          & 0.543          & 0.485          & 0.402          & 0.399          \\ \cline{3-11} 
                          &                                                                              & Avg                                                                          & 0.452          & 0.443          & 0.371          & 0.398          & 0.426          & 0.421          & \textbf{0.281} & 0.327          \\ \hline
\end{tabular}
}
\end{table*}

To verify the effectiveness of the residual connections in the model structure of this paper, ablation experiments were conducted on the two residual connections in Figure \ref{fig:model}, namely, Embedding 1 to the input of the first pair of twin mamba and Embedding 2 to the input of the second pair of twin mamba.
Table \ref{table:residual-channel} presents the results of ablation experiments on residual connections.
In the ``Residual'' column, ``w/o'' indicates the removal of residual connections, while ``Residual'' indicates the presence of residual connections.
As shown in Table \ref{table:residual-channel}, networks with residual connections demonstrate a noticeable improvement in experimental results.

\subsubsection{Ablation of Channel Independence}

To verify the effectiveness of channel independence in the model structure of this paper, ablation experiments were conducted on the Channel Independence module and the Reversed Channel Independence module in Figure \ref{fig:model}.
As shown in Table \ref{table:residual-channel}, in the ``Channel Independence'' column, ``w/o'' indicates the removal of the Channel Independence/Reversed Channel Independence module, while ``Channel Independence'' indicates the presence of the corresponding module.
From the results, it can be seen that networks with the Channel Independence/Reversed Channel Independence module demonstrate a noticeable improvement in experimental results.

\subsubsection{Scalability of \textsf{DTMamba}}

\begin{table*}[t]
\caption{Scalability of \textsf{DTMamba}}
\label{table:scalability}
\resizebox{1.7\expwidths}{!}{
\begin{tabular}{c|c|cc|cc|cc|cc|ll|ll}
\hline
\multirow{2}{*}{Design}  & \multirow{2}{*}{\begin{tabular}[c]{@{}c@{}}Forecast\\ length $S$\end{tabular}} & \multicolumn{2}{c|}{ETTh1}      & \multicolumn{2}{c|}{ETTh2}      & \multicolumn{2}{c|}{ETTm1}      & \multicolumn{2}{c|}{ETTm2}      & \multicolumn{2}{c|}{Weather}                         & \multicolumn{2}{c}{Exchange}                        \\ \cline{3-14} 
                         &                                                                              & MSE            & MAE            & MSE            & MAE            & MSE            & MAE            & MSE            & MAE            & \multicolumn{1}{c}{MSE}   & \multicolumn{1}{c|}{MAE} & \multicolumn{1}{c}{MSE} & \multicolumn{1}{c}{MAE}   \\ \hline
\multirow{5}{*}{DTMamba} & 96                                                                           & 0.386          & 0.399          & 0.290          & 0.340          & 0.325          & 0.360          & 0.177          & 0.259          & 0.171                     & 0.218                    & 0.083                   & 0.201                     \\
                         & 192                                                                          & 0.426          & 0.424          & 0.366          & 0.392          & 0.375          & 0.386          & 0.240          & 0.300          & 0.220                     & 0.257                    & 0.173                   & 0.295                     \\
                         & 336                                                                          & 0.480          & 0.450          & 0.380          & 0.409          & 0.396          & 0.405          & 0.310          & 0.345          & 0.274                     & 0.296                    & 0.346                   & 0.427                     \\
                         & 720                                                                          & 0.484          & 0.470          & 0.416          & 0.437          & 0.454          & 0.442          & 0.395          & 0.394          & 0.349                     & 0.346                    & 0.868                   & 0.698                     \\ \cline{2-14} 
                         & Avg                                                                          & \textbf{0.444} & \textbf{0.435} & \textbf{0.363} & \textbf{0.395} & \textbf{0.388} & \textbf{0.399} & 0.281          & 0.325          & 0.254                     & 0.279                    & 0.368                   & 0.405                     \\ \hline
\multirow{5}{*}{Mamba}   & 96                                                                           & 0.379          & 0.398          & 0.295          & 0.347          & 0.320          & 0.357          & 0.176          & 0.258          & 0.164                     & 0.208                    & 0.085                   & 0.202                     \\
                         & 192                                                                          & 0.440          & 0.426          & 0.364          & 0.390          & 0.373          & 0.388          & 0.241          & 0.301          & 0.208                     & 0.249                    & 0.172                   & 0.295                     \\
                         & 336                                                                          & 0.474          & 0.444          & 0.414          & 0.425          & 0.398          & 0.407          & 0.295          & 0.336          & 0.267                     & 0.290                    & 0.333                   & 0.417                     \\
                         & 720                                                                          & 0.494          & 0.479          & 0.432          & 0.447          & 0.462          & 0.446          & 0.397          & 0.395          & 0.346                     & 0.342                    & 0.882                   & 0.703                     \\ \cline{2-14} 
                         & Avg                                                                          & 0.447          & 0.437          & 0.376          & 0.409          & \textbf{0.388} & 0.400          & \textbf{0.277} & \textbf{0.323} & \textbf{0.246}            & \textbf{0.272}           & 0.368                   & 0.404                     \\ \hline
\multirow{5}{*}{DMamba}  & 96                                                                           & 0.377          & 0.396          & 0.292          & 0.343          & 0.326          & 0.363          & 0.180          & 0.258          & 0.179                     & 0.224                    & 0.084                   & 0.201                     \\
                         & 192                                                                          & 0.433          & 0.426          & 0.367          & 0.391          & 0.373          & 0.386          & 0.239          & 0.299          & 0.222                     & 0.259                    & 0.174                   & 0.296                     \\
                         & 336                                                                          & 0.490          & 0.456          & 0.415          & 0.428          & 0.442          & 0.408          & 0.297          & 0.338          & 0.278                     & 0.299                    & 0.345                   & 0.424                     \\
                         & 720                                                                          & 0.524          & 0.492          & 0.426          & 0.443          & 0.460          & 0.443          & 0.397          & 0.395          & 0.349                     & 0.345                    & 0.844                   & 0.689                     \\ \cline{2-14} 
                         & Avg                                                                          & 0.455          & 0.443          & 0.375          & 0.401          & 0.400          & 0.400          & 0.278          & \textbf{0.323} & 0.257                     & 0.282                    & \textbf{0.362}          & \textbf{0.403}            \\ \hline
\multirow{5}{*}{TMamba}  & 96                                                                           & 0.380          & 0.396          & 0.289          & 0.342          & 0.318          & 0.356          & 0.178          & 0.259          & 0.165                     & 0.212                    & 0.084                   & 0.200                     \\
                         & 192                                                                          & 0.441          & 0.430          & 0.373          & 0.392          & 0.369          & 0.385          & 0.242          & 0.303          & \multicolumn{1}{c}{0.212} & 0.253                    & 0.176                   & \multicolumn{1}{c}{0.297} \\
                         & 336                                                                          & 0.483          & 0.449          & 0.411          & 0.425          & 0.405          & 0.409          & 0.303          & 0.341          & \multicolumn{1}{c}{0.267} & 0.291                    & 0.347                   & \multicolumn{1}{c}{0.425} \\
                         & 720                                                                          & 0.490          & 0.475          & 0.423          & 0.441          & 0.463          & 0.444          & 0.399          & 0.398          & \multicolumn{1}{c}{0.345} & 0.343                    & 0.940                   & \multicolumn{1}{c}{0.714} \\ \cline{2-14} 
                         & Avg                                                                          & 0.449          & 0.438          & 0.374          & 0.400          & 0.389          & \textbf{0.399} & 0.281          & 0.325          & \multicolumn{1}{c}{0.247} & 0.275                    & 0.387                   & \multicolumn{1}{c}{0.409} \\ \hline
\end{tabular}
}
\end{table*}

To evaluate the effectiveness and scalability of the proposed \textsf{TMamba Block}, we conduct module replacement experiments on the ETTh1, ETTh2, ETTm1, and ETTm2 datasets.
As shown in Table \ref{table:scalability}, in the ``Design'' column, ``DTMamba'' represents the two TMamba blocks used in this paper, ``Mamba'' represents the use of a single Mamba model to replace two TMamba blocks, ``DMamba'' represents the use of two TMamba blocks, but each TMamba block only uses one Mamba, and ``TMamba'' represents the use of only one TMamba block proposed in this paper.
From the experimental results, it can be observed that the proposed \textsf{DTMamba} has the best performance.
This structure is capable of better learning relevant features and temporal patterns in the time series data.

\subsubsection{Sensitivity of Dropouts}

\begin{figure}[t]
  \centering
  \includegraphics[width=\expwidths]{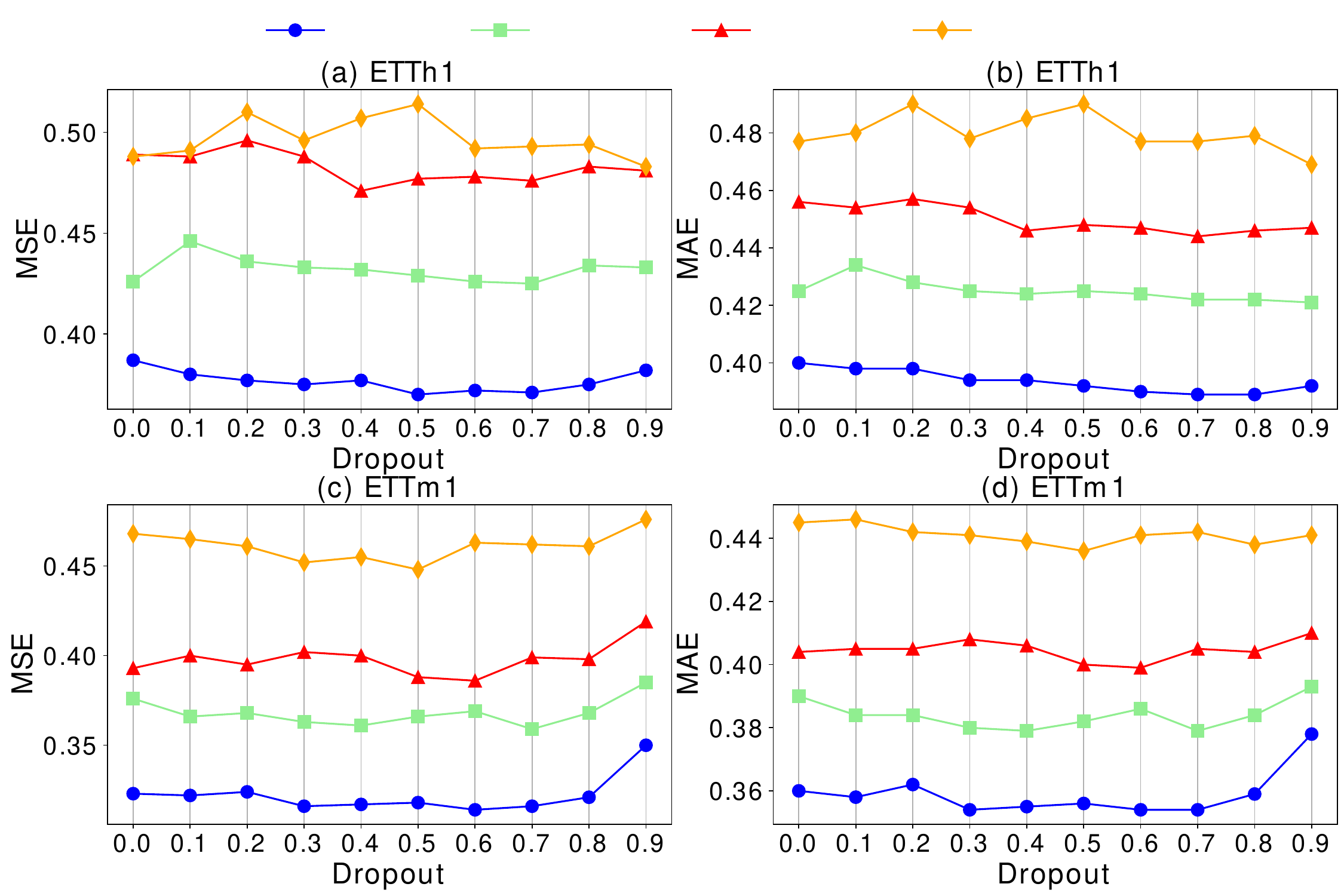}
  \caption{Sensitivity of Dropouts}%
  \label{exp:dropout}
\end{figure}

As shown in Figure \ref{exp:dropout}, except for the cases with $\mathit{Dropout}=0.2/0.4/0.5$ for a prediction length of $\mathit{S}=720$ in Figure \ref{exp:dropout}(a)(b), and $\mathit{Dropout}=0.9$ in Figure \ref{exp:dropout}(c)(d), the overall experimental results did not show significant fluctuations with changes in Dropout.
Additionally, in most cases, results with $\mathit{Dropout}>0$ outperform $\mathit{Dropout}=0$ (i.e., adding Dropout).
However, increasing Dropout leads to higher computational costs during training.
Therefore, for the majority of experiments, we chose $\mathit{Dropout}=0.05$, which offer a good balance between effectiveness and computational efficiency.

\subsubsection{Sensitivity of Linear's Parameters}
\label{sect:linear}

\begin{figure}[t]
  \centering
  \includegraphics[width=\expwidths]{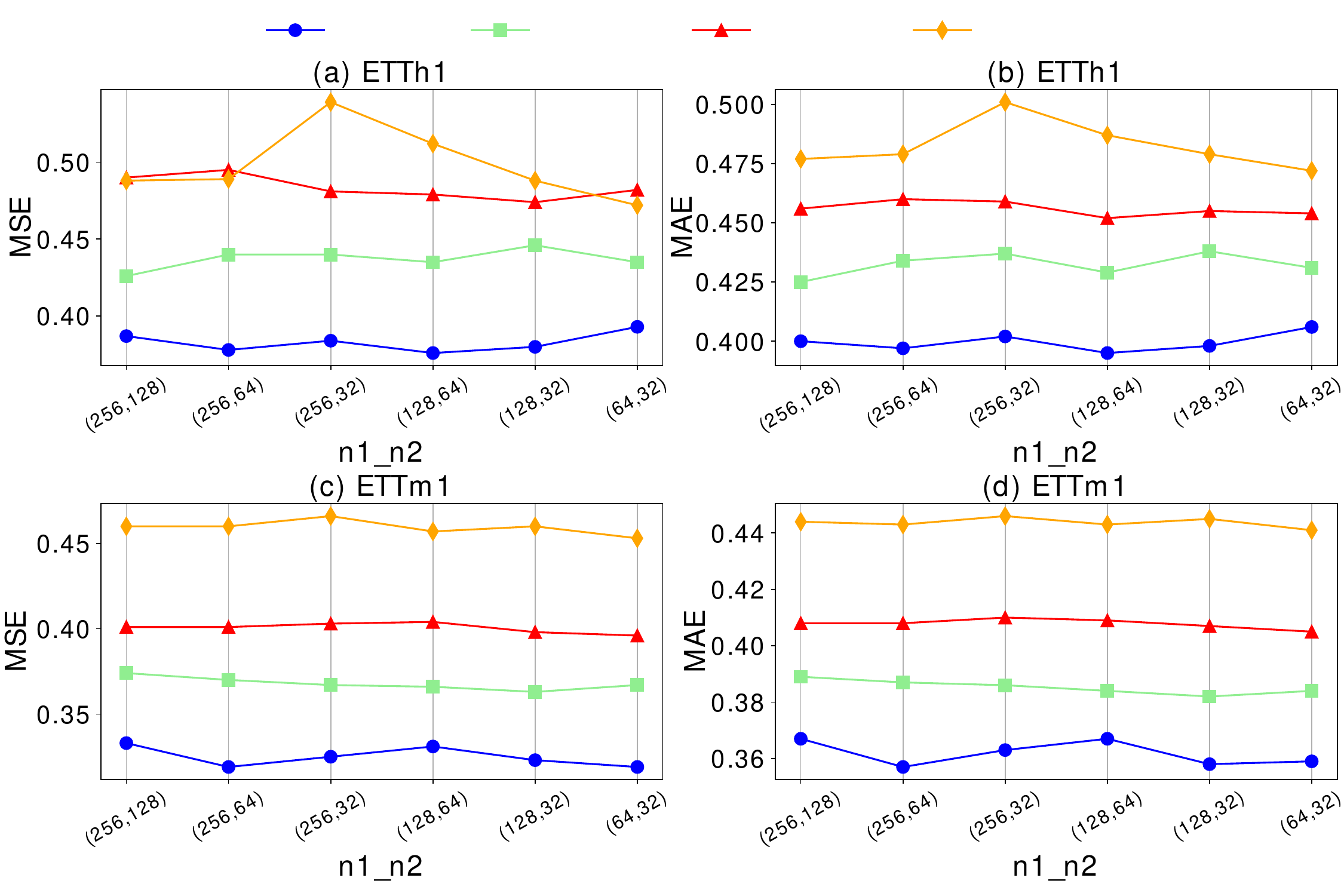}
  \caption{Sensitivity of Linear's Parameters}%
  \label{exp:n1-n2}
\end{figure}

As shown in Figure \ref{fig:model}, our proposed network structure includes two linear layers serving as Embedding layers, namely Embedding 1 and Embedding 2.
Since the baseline models, \textsf{DLinear} and \textsf{RLinear}, have demonstrated the crucial role of linear mapping in previous long-term time series prediction works, to understand the performance impact of MLP on our network, we consider different combinations of linear dimensions.
As depicted in Figure \ref{exp:n1-n2}, under different combinations of linear dimensions, the performance of \textsf{DTMamba} on ETTh1 and ETTm1 does not depend on the choice of MLP.
Additionally, for multidimensional time series data with smaller dimensions, selecting smaller combinations of linear dimensions yields better results.

\subsubsection{Sensitive of Mamba Dimension Expansion Factor}

\begin{figure}[t]
  \centering
  \includegraphics[width=\expwidths]{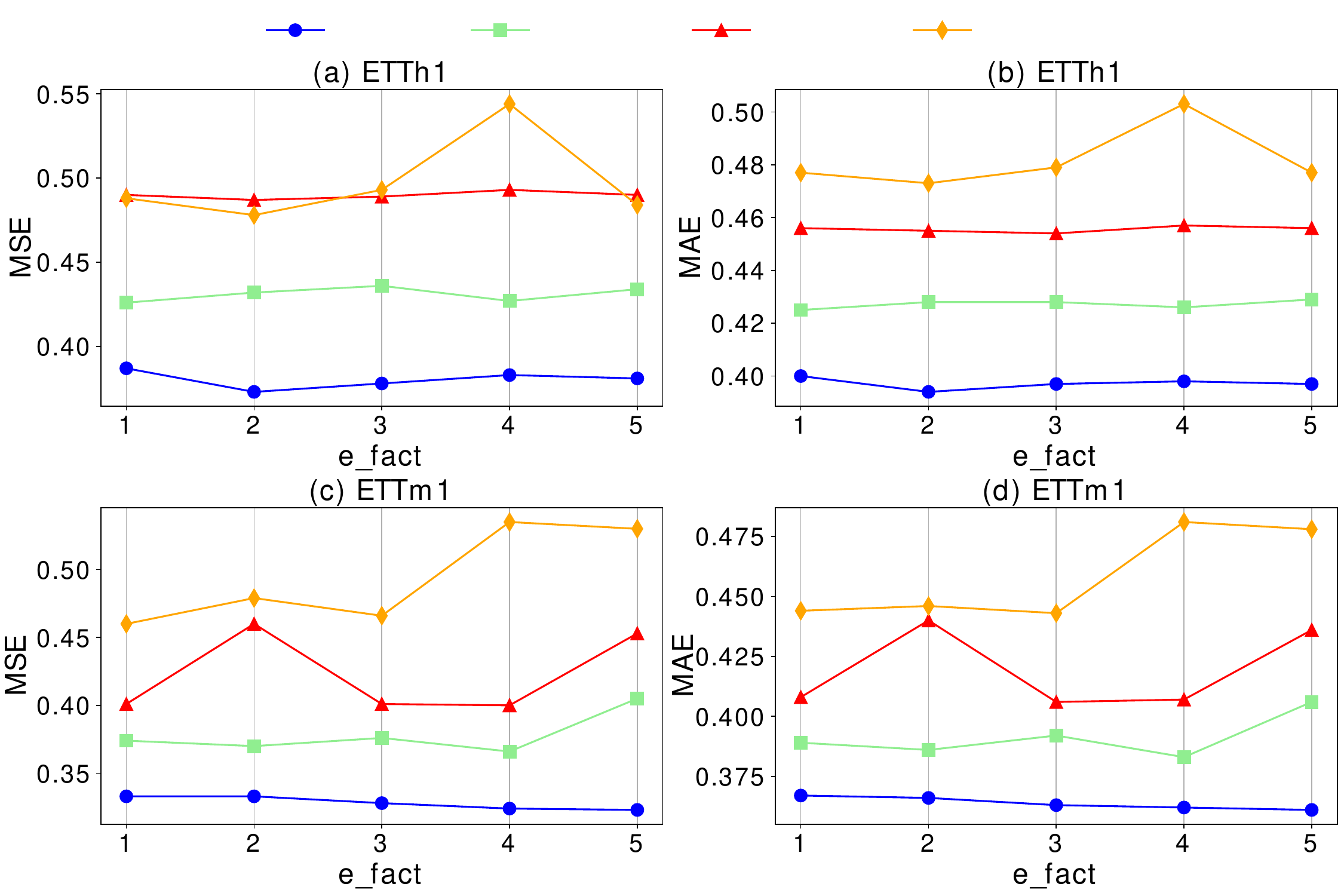}
  \caption{Sensitive of Mamba Dimension Expansion Factor}%
  \label{exp:e-fact}
\end{figure}

We conduct sensitivity experiments on the dimension expansion factor parameter ($\mathit{e\_fact}$) of \textsf{Mamba}.
As shown in Figure \ref{exp:e-fact}, increasing the dimension expansion factor sometimes leads to a deterioration in experimental results, along with significant time and memory consumption. 
Therefore, we chose $\mathit{e\_fact}=1$ (the default value for Mamba) for all experiments, as it offer better performance with lower memory and time costs.

\subsubsection{Sensitive of State Expansion Factor of Mambas}

\begin{figure}[t]
  \centering
  \includegraphics[width=\expwidths]{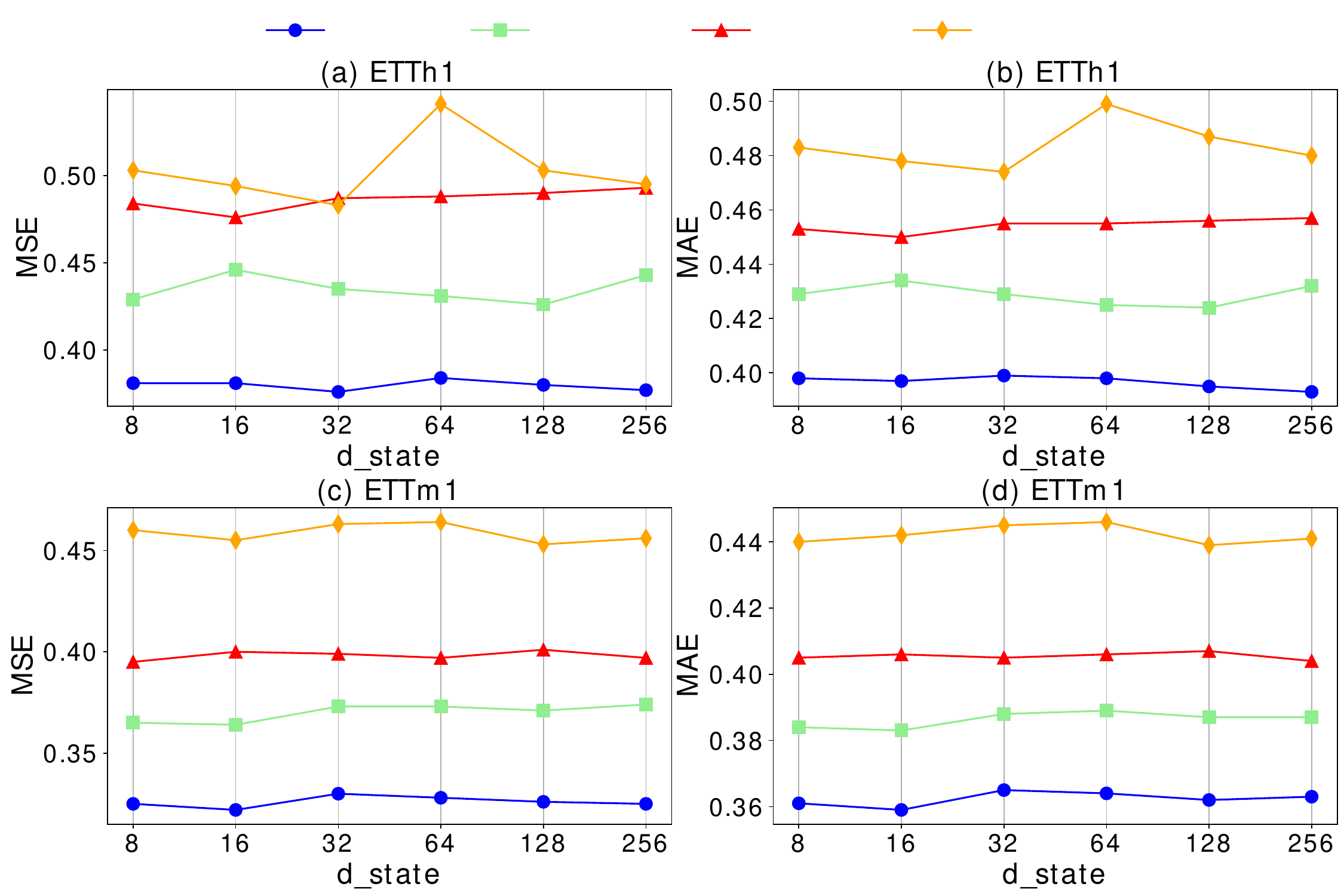}
  \caption{Sensitive of State Expansion Factor of Mambas}%
  \label{exp:d-state}
\end{figure}

Similarly, we conduct sensitivity experiments on another key parameter of \textsf{Mamba}, the state expansion factor of the \textsf{SSM} ($\mathit{d\_state}$).
As shown in Figure \ref{exp:d-state}, except for the case of $\mathit{d\_state}=64$ for a prediction length of $\mathit{S}=720$ in Figure \ref{exp:d-state}(a)(b), the experimental results did not show significant fluctuations with changes in $\mathit{d\_state}$.
Overall, with $\mathit{d\_state}=256$, the experimental results were stable and effective.
Therefore, we set $\mathit{d\_state}=256$ (the default value for Mamba) for all experiments.

\subsubsection{Sensitive of Mambas' Local Convolutional Width}

\begin{figure}[t]
  \centering
  \includegraphics[width=\expwidths]{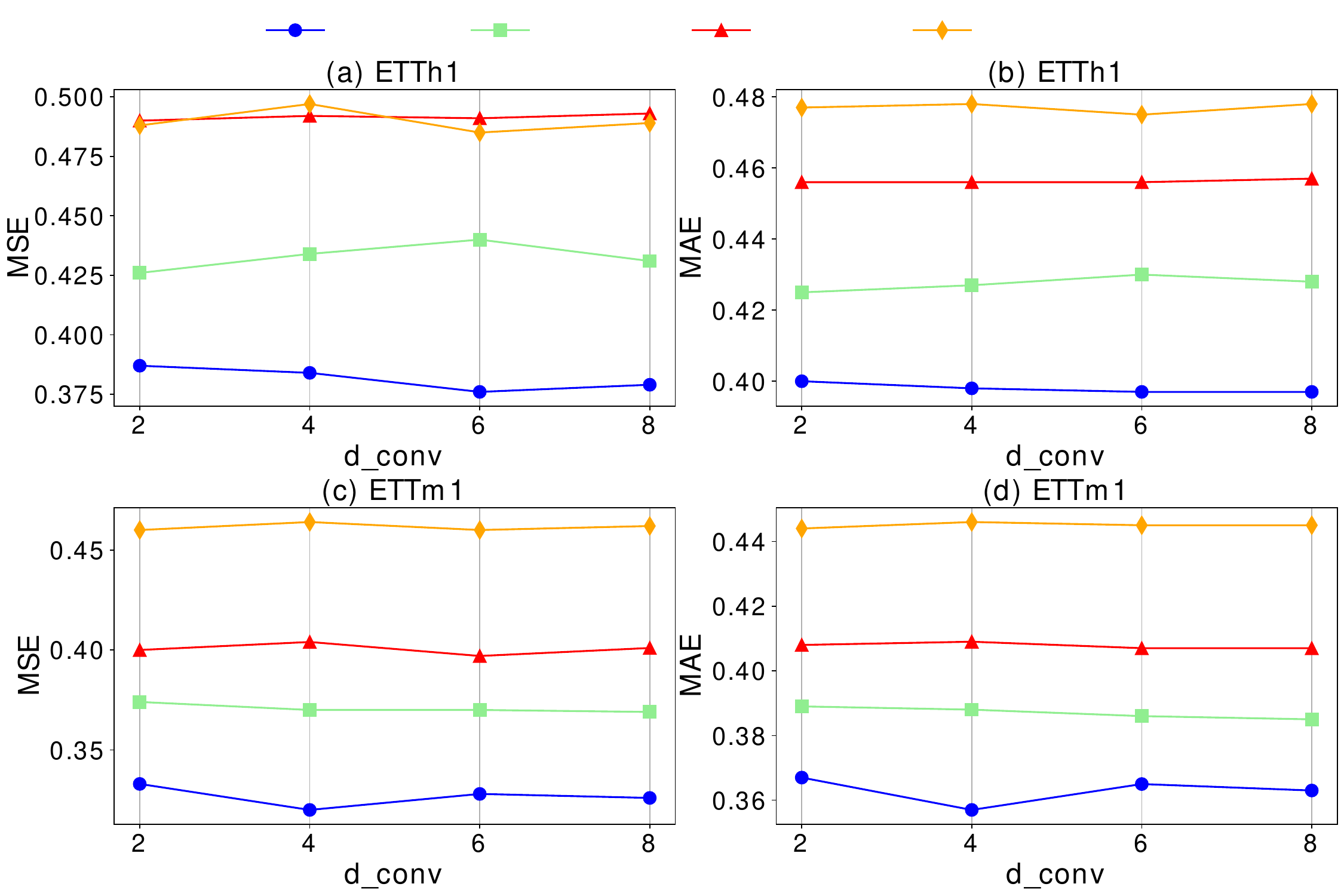}
  \caption{Sensitive of Mambas' Local Convolutional Width}%
  \label{exp:d-conv}
\end{figure}

Finally, we conduct sensitivity experiments on the width parameter of the local convolutional kernel in \textsf{Mamba} ($\mathit{d\_conv}$).
As shown in Figure \ref{exp:d-conv}, the experimental results did not exhibit significant fluctuations with changes in $\mathit{d\_conv}$.
Moreover, increasing the width of the local convolution kernel did not lead to improvements in experimental results.
Therefore, we set $\mathit{d\_conv}=2$ (the default value for Mamba) for all experiments.

%% file: mamba.bbl

\begin{thebibliography}{40}


\ifx \showCODEN    \undefined \def \showCODEN     #1{\unskip}     \fi
\ifx \showDOI      \undefined \def \showDOI       #1{#1}\fi
\ifx \showISBNx    \undefined \def \showISBNx     #1{\unskip}     \fi
\ifx \showISBNxiii \undefined \def \showISBNxiii  #1{\unskip}     \fi
\ifx \showISSN     \undefined \def \showISSN      #1{\unskip}     \fi
\ifx \showLCCN     \undefined \def \showLCCN      #1{\unskip}     \fi
\ifx \shownote     \undefined \def \shownote      #1{#1}          \fi
\ifx \showarticletitle \undefined \def \showarticletitle #1{#1}   \fi
\ifx \showURL      \undefined \def \showURL       {\relax}        \fi
\providecommand\bibfield[2]{#2}
\providecommand\bibinfo[2]{#2}
\providecommand\natexlab[1]{#1}
\providecommand\showeprint[2][]{arXiv:#2}

\bibitem[Behrouz and Hashemi(2024)]%
        {DBLP:journals/corr/abs-2402-08678}
\bibfield{author}{\bibinfo{person}{Ali Behrouz} {and} \bibinfo{person}{Farnoosh
  Hashemi}.} \bibinfo{year}{2024}\natexlab{}.
\newblock \showarticletitle{Graph Mamba: Towards Learning on Graphs with State
  Space Models}.
\newblock \bibinfo{journal}{\emph{CoRR}}  \bibinfo{volume}{abs/2402.08678}
  (\bibinfo{year}{2024}).
\newblock


\bibitem[Box and Jenkins(1968)]%
        {box1968some}
\bibfield{author}{\bibinfo{person}{George~EP Box} {and}
  \bibinfo{person}{Gwilym~M Jenkins}.} \bibinfo{year}{1968}\natexlab{}.
\newblock \showarticletitle{Some recent advances in forecasting and control}.
\newblock \bibinfo{journal}{\emph{Journal of the Royal Statistical Society.
  Series C (Applied Statistics)}} \bibinfo{volume}{17}, \bibinfo{number}{2}
  (\bibinfo{year}{1968}), \bibinfo{pages}{91--109}.
\newblock


\bibitem[Chen et~al\mbox{.}(2023)]%
        {DBLP:journals/corr/abs-2303-06053}
\bibfield{author}{\bibinfo{person}{Si{-}An Chen}, \bibinfo{person}{Chun{-}Liang
  Li}, \bibinfo{person}{Nate Yoder}, \bibinfo{person}{Sercan~{\"{O}}. Arik},
  {and} \bibinfo{person}{Tomas Pfister}.} \bibinfo{year}{2023}\natexlab{}.
\newblock \showarticletitle{TSMixer: An all-MLP Architecture for Time Series
  Forecasting}.
\newblock \bibinfo{journal}{\emph{CoRR}}  \bibinfo{volume}{abs/2303.06053}
  (\bibinfo{year}{2023}).
\newblock


\bibitem[Das et~al\mbox{.}(2023)]%
        {DBLP:journals/corr/abs-2304-08424}
\bibfield{author}{\bibinfo{person}{Abhimanyu Das}, \bibinfo{person}{Weihao
  Kong}, \bibinfo{person}{Andrew Leach}, \bibinfo{person}{Shaan Mathur},
  \bibinfo{person}{Rajat Sen}, {and} \bibinfo{person}{Rose Yu}.}
  \bibinfo{year}{2023}\natexlab{}.
\newblock \showarticletitle{Long-term Forecasting with TiDE: Time-series Dense
  Encoder}.
\newblock \bibinfo{journal}{\emph{CoRR}}  \bibinfo{volume}{abs/2304.08424}
  (\bibinfo{year}{2023}).
\newblock


\bibitem[Das et~al\mbox{.}(2004)]%
        {das2004mean}
\bibfield{author}{\bibinfo{person}{Kalyan Das}, \bibinfo{person}{Jiming Jiang},
  {and} \bibinfo{person}{JNK Rao}.} \bibinfo{year}{2004}\natexlab{}.
\newblock \showarticletitle{Mean squared error of empirical predictor}.
\newblock  (\bibinfo{year}{2004}).
\newblock


\bibitem[Elman(1990)]%
        {DBLP:journals/cogsci/Elman90}
\bibfield{author}{\bibinfo{person}{Jeffrey~L. Elman}.}
  \bibinfo{year}{1990}\natexlab{}.
\newblock \showarticletitle{Finding Structure in Time}.
\newblock \bibinfo{journal}{\emph{Cogn. Sci.}} \bibinfo{volume}{14},
  \bibinfo{number}{2} (\bibinfo{year}{1990}), \bibinfo{pages}{179--211}.
\newblock


\bibitem[Gu and Dao(2023)]%
        {DBLP:journals/corr/abs-2312-00752}
\bibfield{author}{\bibinfo{person}{Albert Gu} {and} \bibinfo{person}{Tri Dao}.}
  \bibinfo{year}{2023}\natexlab{}.
\newblock \showarticletitle{Mamba: Linear-Time Sequence Modeling with Selective
  State Spaces}.
\newblock \bibinfo{journal}{\emph{CoRR}}  \bibinfo{volume}{abs/2312.00752}
  (\bibinfo{year}{2023}).
\newblock


\bibitem[Gu et~al\mbox{.}(2022)]%
        {DBLP:conf/iclr/GuGR22}
\bibfield{author}{\bibinfo{person}{Albert Gu}, \bibinfo{person}{Karan Goel},
  {and} \bibinfo{person}{Christopher R{\'{e}}}.}
  \bibinfo{year}{2022}\natexlab{}.
\newblock \showarticletitle{Efficiently Modeling Long Sequences with Structured
  State Spaces}. In \bibinfo{booktitle}{\emph{{ICLR}}}.
  \bibinfo{publisher}{OpenReview.net}.
\newblock


\bibitem[He et~al\mbox{.}(2016a)]%
        {DBLP:conf/cvpr/HeZRS16}
\bibfield{author}{\bibinfo{person}{Kaiming He}, \bibinfo{person}{Xiangyu
  Zhang}, \bibinfo{person}{Shaoqing Ren}, {and} \bibinfo{person}{Jian Sun}.}
  \bibinfo{year}{2016}\natexlab{a}.
\newblock \showarticletitle{Deep Residual Learning for Image Recognition}. In
  \bibinfo{booktitle}{\emph{{CVPR}}}. \bibinfo{publisher}{{IEEE} Computer
  Society}, \bibinfo{pages}{770--778}.
\newblock


\bibitem[He et~al\mbox{.}(2016b)]%
        {DBLP:conf/eccv/HeZRS16}
\bibfield{author}{\bibinfo{person}{Kaiming He}, \bibinfo{person}{Xiangyu
  Zhang}, \bibinfo{person}{Shaoqing Ren}, {and} \bibinfo{person}{Jian Sun}.}
  \bibinfo{year}{2016}\natexlab{b}.
\newblock \showarticletitle{Identity Mappings in Deep Residual Networks}. In
  \bibinfo{booktitle}{\emph{{ECCV} {(4)}}} \emph{(\bibinfo{series}{Lecture
  Notes in Computer Science}, Vol.~\bibinfo{volume}{9908})}.
  \bibinfo{publisher}{Springer}, \bibinfo{pages}{630--645}.
\newblock


\bibitem[Hewage et~al\mbox{.}(2020)]%
        {DBLP:journals/soco/HewageBTPGPL20}
\bibfield{author}{\bibinfo{person}{Pradeep Hewage}, \bibinfo{person}{Ardhendu
  Behera}, \bibinfo{person}{Marcello Trovati}, \bibinfo{person}{Ella Pereira},
  \bibinfo{person}{Morteza Ghahremani}, \bibinfo{person}{Francesco Palmieri},
  {and} \bibinfo{person}{Yonghuai Liu}.} \bibinfo{year}{2020}\natexlab{}.
\newblock \showarticletitle{Temporal convolutional neural {(TCN)} network for
  an effective weather forecasting using time-series data from the local
  weather station}.
\newblock \bibinfo{journal}{\emph{Soft Comput.}} \bibinfo{volume}{24},
  \bibinfo{number}{21} (\bibinfo{year}{2020}), \bibinfo{pages}{16453--16482}.
\newblock


\bibitem[Hinton et~al\mbox{.}(2012)]%
        {DBLP:journals/corr/abs-1207-0580}
\bibfield{author}{\bibinfo{person}{Geoffrey~E. Hinton}, \bibinfo{person}{Nitish
  Srivastava}, \bibinfo{person}{Alex Krizhevsky}, \bibinfo{person}{Ilya
  Sutskever}, {and} \bibinfo{person}{Ruslan Salakhutdinov}.}
  \bibinfo{year}{2012}\natexlab{}.
\newblock \showarticletitle{Improving neural networks by preventing
  co-adaptation of feature detectors}.
\newblock \bibinfo{journal}{\emph{CoRR}}  \bibinfo{volume}{abs/1207.0580}
  (\bibinfo{year}{2012}).
\newblock


\bibitem[Karevan and Suykens(2020)]%
        {DBLP:journals/nn/KarevanS20}
\bibfield{author}{\bibinfo{person}{Zahra Karevan} {and} \bibinfo{person}{Johan
  A.~K. Suykens}.} \bibinfo{year}{2020}\natexlab{}.
\newblock \showarticletitle{Transductive {LSTM} for time-series prediction: An
  application to weather forecasting}.
\newblock \bibinfo{journal}{\emph{Neural Networks}}  \bibinfo{volume}{125}
  (\bibinfo{year}{2020}), \bibinfo{pages}{1--9}.
\newblock


\bibitem[Kim et~al\mbox{.}(2022)]%
        {DBLP:conf/iclr/KimKTPCC22}
\bibfield{author}{\bibinfo{person}{Taesung Kim}, \bibinfo{person}{Jinhee Kim},
  \bibinfo{person}{Yunwon Tae}, \bibinfo{person}{Cheonbok Park},
  \bibinfo{person}{Jang{-}Ho Choi}, {and} \bibinfo{person}{Jaegul Choo}.}
  \bibinfo{year}{2022}\natexlab{}.
\newblock \showarticletitle{Reversible Instance Normalization for Accurate
  Time-Series Forecasting against Distribution Shift}. In
  \bibinfo{booktitle}{\emph{{ICLR}}}. \bibinfo{publisher}{OpenReview.net}.
\newblock


\bibitem[Kim(2014)]%
        {DBLP:conf/emnlp/Kim14}
\bibfield{author}{\bibinfo{person}{Yoon Kim}.} \bibinfo{year}{2014}\natexlab{}.
\newblock \showarticletitle{Convolutional Neural Networks for Sentence
  Classification}. In \bibinfo{booktitle}{\emph{{EMNLP}}}.
  \bibinfo{publisher}{{ACL}}, \bibinfo{pages}{1746--1751}.
\newblock


\bibitem[Kingma and Ba(2015)]%
        {DBLP:journals/corr/KingmaB14}
\bibfield{author}{\bibinfo{person}{Diederik~P. Kingma} {and}
  \bibinfo{person}{Jimmy Ba}.} \bibinfo{year}{2015}\natexlab{}.
\newblock \showarticletitle{Adam: {A} Method for Stochastic Optimization}. In
  \bibinfo{booktitle}{\emph{{ICLR} (Poster)}}.
\newblock


\bibitem[Lai et~al\mbox{.}(2018)]%
        {DBLP:conf/sigir/LaiCYL18}
\bibfield{author}{\bibinfo{person}{Guokun Lai}, \bibinfo{person}{Wei{-}Cheng
  Chang}, \bibinfo{person}{Yiming Yang}, {and} \bibinfo{person}{Hanxiao Liu}.}
  \bibinfo{year}{2018}\natexlab{}.
\newblock \showarticletitle{Modeling Long- and Short-Term Temporal Patterns
  with Deep Neural Networks}. In \bibinfo{booktitle}{\emph{{SIGIR}}}.
  \bibinfo{publisher}{{ACM}}, \bibinfo{pages}{95--104}.
\newblock


\bibitem[Li et~al\mbox{.}(2024)]%
        {DBLP:journals/corr/abs-2403-06977}
\bibfield{author}{\bibinfo{person}{Kunchang Li}, \bibinfo{person}{Xinhao Li},
  \bibinfo{person}{Yi Wang}, \bibinfo{person}{Yinan He}, \bibinfo{person}{Yali
  Wang}, \bibinfo{person}{Limin Wang}, {and} \bibinfo{person}{Yu Qiao}.}
  \bibinfo{year}{2024}\natexlab{}.
\newblock \showarticletitle{VideoMamba: State Space Model for Efficient Video
  Understanding}.
\newblock \bibinfo{journal}{\emph{CoRR}}  \bibinfo{volume}{abs/2403.06977}
  (\bibinfo{year}{2024}).
\newblock


\bibitem[Li et~al\mbox{.}(2023)]%
        {DBLP:journals/corr/abs-2305-10721}
\bibfield{author}{\bibinfo{person}{Zhe Li}, \bibinfo{person}{Shiyi Qi},
  \bibinfo{person}{Yiduo Li}, {and} \bibinfo{person}{Zenglin Xu}.}
  \bibinfo{year}{2023}\natexlab{}.
\newblock \showarticletitle{Revisiting Long-term Time Series Forecasting: An
  Investigation on Linear Mapping}.
\newblock \bibinfo{journal}{\emph{CoRR}}  \bibinfo{volume}{abs/2305.10721}
  (\bibinfo{year}{2023}).
\newblock


\bibitem[Lieber et~al\mbox{.}(2024)]%
        {DBLP:journals/corr/abs-2403-19887}
\bibfield{author}{\bibinfo{person}{Opher Lieber}, \bibinfo{person}{Barak Lenz},
  \bibinfo{person}{Hofit Bata}, \bibinfo{person}{Gal Cohen},
  \bibinfo{person}{Jhonathan Osin}, \bibinfo{person}{Itay Dalmedigos},
  \bibinfo{person}{Erez Safahi}, \bibinfo{person}{Shaked Meirom},
  \bibinfo{person}{Yonatan Belinkov}, \bibinfo{person}{Shai Shalev{-}Shwartz},
  \bibinfo{person}{Omri Abend}, \bibinfo{person}{Raz Alon},
  \bibinfo{person}{Tomer Asida}, \bibinfo{person}{Amir Bergman},
  \bibinfo{person}{Roman Glozman}, \bibinfo{person}{Michael Gokhman},
  \bibinfo{person}{Avashalom Manevich}, \bibinfo{person}{Nir Ratner},
  \bibinfo{person}{Noam Rozen}, \bibinfo{person}{Erez Shwartz},
  \bibinfo{person}{Mor Zusman}, {and} \bibinfo{person}{Yoav Shoham}.}
  \bibinfo{year}{2024}\natexlab{}.
\newblock \showarticletitle{Jamba: {A} Hybrid Transformer-Mamba Language
  Model}.
\newblock \bibinfo{journal}{\emph{CoRR}}  \bibinfo{volume}{abs/2403.19887}
  (\bibinfo{year}{2024}).
\newblock


\bibitem[Liu et~al\mbox{.}(2022b)]%
        {DBLP:conf/nips/LiuZCXLM022}
\bibfield{author}{\bibinfo{person}{Minhao Liu}, \bibinfo{person}{Ailing Zeng},
  \bibinfo{person}{Muxi Chen}, \bibinfo{person}{Zhijian Xu},
  \bibinfo{person}{Qiuxia Lai}, \bibinfo{person}{Lingna Ma}, {and}
  \bibinfo{person}{Qiang Xu}.} \bibinfo{year}{2022}\natexlab{b}.
\newblock \showarticletitle{SCINet: Time Series Modeling and Forecasting with
  Sample Convolution and Interaction}. In \bibinfo{booktitle}{\emph{NeurIPS}}.
\newblock


\bibitem[Liu et~al\mbox{.}(2023)]%
        {DBLP:journals/corr/abs-2310-06625}
\bibfield{author}{\bibinfo{person}{Yong Liu}, \bibinfo{person}{Tengge Hu},
  \bibinfo{person}{Haoran Zhang}, \bibinfo{person}{Haixu Wu},
  \bibinfo{person}{Shiyu Wang}, \bibinfo{person}{Lintao Ma}, {and}
  \bibinfo{person}{Mingsheng Long}.} \bibinfo{year}{2023}\natexlab{}.
\newblock \showarticletitle{iTransformer: Inverted Transformers Are Effective
  for Time Series Forecasting}.
\newblock \bibinfo{journal}{\emph{CoRR}}  \bibinfo{volume}{abs/2310.06625}
  (\bibinfo{year}{2023}).
\newblock


\bibitem[Liu et~al\mbox{.}(2022a)]%
        {liu2022non}
\bibfield{author}{\bibinfo{person}{Yong Liu}, \bibinfo{person}{Haixu Wu},
  \bibinfo{person}{Jianmin Wang}, {and} \bibinfo{person}{Mingsheng Long}.}
  \bibinfo{year}{2022}\natexlab{a}.
\newblock \showarticletitle{Non-stationary transformers: Exploring the
  stationarity in time series forecasting}.
\newblock \bibinfo{journal}{\emph{Advances in Neural Information Processing
  Systems}}  \bibinfo{volume}{35} (\bibinfo{year}{2022}),
  \bibinfo{pages}{9881--9893}.
\newblock


\bibitem[Ma et~al\mbox{.}(2024)]%
        {DBLP:journals/corr/abs-2401-04722}
\bibfield{author}{\bibinfo{person}{Jun Ma}, \bibinfo{person}{Feifei Li}, {and}
  \bibinfo{person}{Bo Wang}.} \bibinfo{year}{2024}\natexlab{}.
\newblock \showarticletitle{U-Mamba: Enhancing Long-range Dependency for
  Biomedical Image Segmentation}.
\newblock \bibinfo{journal}{\emph{CoRR}}  \bibinfo{volume}{abs/2401.04722}
  (\bibinfo{year}{2024}).
\newblock


\bibitem[Madan and Sarathi(2018)]%
        {DBLP:conf/ic3/MadanM18}
\bibfield{author}{\bibinfo{person}{Rishabh Madan} {and}
  \bibinfo{person}{Mangipudi~Partha Sarathi}.} \bibinfo{year}{2018}\natexlab{}.
\newblock \showarticletitle{Predicting Computer Network Traffic: {A} Time
  Series Forecasting Approach Using DWT, {ARIMA} and {RNN}}. In
  \bibinfo{booktitle}{\emph{{IC3}}}. \bibinfo{publisher}{{IEEE} Computer
  Society}, \bibinfo{pages}{1--5}.
\newblock


\bibitem[Morid et~al\mbox{.}(2023)]%
        {DBLP:journals/tmis/MoridSD23}
\bibfield{author}{\bibinfo{person}{Mohammad~Amin Morid},
  \bibinfo{person}{Olivia R.~Liu Sheng}, {and} \bibinfo{person}{Joseph
  Dunbar}.} \bibinfo{year}{2023}\natexlab{}.
\newblock \showarticletitle{Time Series Prediction Using Deep Learning Methods
  in Healthcare}.
\newblock \bibinfo{journal}{\emph{{ACM} Trans. Manag. Inf. Syst.}}
  \bibinfo{volume}{14}, \bibinfo{number}{1} (\bibinfo{year}{2023}),
  \bibinfo{pages}{2:1--2:29}.
\newblock


\bibitem[Nie et~al\mbox{.}(2023)]%
        {DBLP:conf/iclr/NieNSK23}
\bibfield{author}{\bibinfo{person}{Yuqi Nie}, \bibinfo{person}{Nam~H. Nguyen},
  \bibinfo{person}{Phanwadee Sinthong}, {and} \bibinfo{person}{Jayant
  Kalagnanam}.} \bibinfo{year}{2023}\natexlab{}.
\newblock \showarticletitle{A Time Series is Worth 64 Words: Long-term
  Forecasting with Transformers}. In \bibinfo{booktitle}{\emph{{ICLR}}}.
  \bibinfo{publisher}{OpenReview.net}.
\newblock


\bibitem[Pacella and Papadia(2021)]%
        {pacella2021evaluation}
\bibfield{author}{\bibinfo{person}{Massimo Pacella} {and}
  \bibinfo{person}{Gabriele Papadia}.} \bibinfo{year}{2021}\natexlab{}.
\newblock \showarticletitle{Evaluation of deep learning with long short-term
  memory networks for time series forecasting in supply chain management}.
\newblock \bibinfo{journal}{\emph{Procedia CIRP}}  \bibinfo{volume}{99}
  (\bibinfo{year}{2021}), \bibinfo{pages}{604--609}.
\newblock


\bibitem[Paszke et~al\mbox{.}(2019)]%
        {DBLP:conf/nips/PaszkeGMLBCKLGA19}
\bibfield{author}{\bibinfo{person}{Adam Paszke}, \bibinfo{person}{Sam Gross},
  \bibinfo{person}{Francisco Massa}, \bibinfo{person}{Adam Lerer},
  \bibinfo{person}{James Bradbury}, \bibinfo{person}{Gregory Chanan},
  \bibinfo{person}{Trevor Killeen}, \bibinfo{person}{Zeming Lin},
  \bibinfo{person}{Natalia Gimelshein}, \bibinfo{person}{Luca Antiga},
  \bibinfo{person}{Alban Desmaison}, \bibinfo{person}{Andreas K{\"{o}}pf},
  \bibinfo{person}{Edward~Z. Yang}, \bibinfo{person}{Zachary DeVito},
  \bibinfo{person}{Martin Raison}, \bibinfo{person}{Alykhan Tejani},
  \bibinfo{person}{Sasank Chilamkurthy}, \bibinfo{person}{Benoit Steiner},
  \bibinfo{person}{Lu Fang}, \bibinfo{person}{Junjie Bai}, {and}
  \bibinfo{person}{Soumith Chintala}.} \bibinfo{year}{2019}\natexlab{}.
\newblock \showarticletitle{PyTorch: An Imperative Style, High-Performance Deep
  Learning Library}. In \bibinfo{booktitle}{\emph{NeurIPS}}.
  \bibinfo{pages}{8024--8035}.
\newblock


\bibitem[Toner and Darlow(2024)]%
        {DBLP:journals/corr/abs-2403-14587}
\bibfield{author}{\bibinfo{person}{William Toner} {and} \bibinfo{person}{Luke
  Darlow}.} \bibinfo{year}{2024}\natexlab{}.
\newblock \showarticletitle{An Analysis of Linear Time Series Forecasting
  Models}.
\newblock \bibinfo{journal}{\emph{CoRR}}  \bibinfo{volume}{abs/2403.14587}
  (\bibinfo{year}{2024}).
\newblock


\bibitem[Vaswani et~al\mbox{.}(2017)]%
        {DBLP:conf/nips/VaswaniSPUJGKP17}
\bibfield{author}{\bibinfo{person}{Ashish Vaswani}, \bibinfo{person}{Noam
  Shazeer}, \bibinfo{person}{Niki Parmar}, \bibinfo{person}{Jakob Uszkoreit},
  \bibinfo{person}{Llion Jones}, \bibinfo{person}{Aidan~N. Gomez},
  \bibinfo{person}{Lukasz Kaiser}, {and} \bibinfo{person}{Illia Polosukhin}.}
  \bibinfo{year}{2017}\natexlab{}.
\newblock \showarticletitle{Attention is All you Need}. In
  \bibinfo{booktitle}{\emph{{NIPS}}}. \bibinfo{pages}{5998--6008}.
\newblock


\bibitem[Wright et~al\mbox{.}(1986)]%
        {wright1986evaluation}
\bibfield{author}{\bibinfo{person}{David~J Wright}, \bibinfo{person}{G Capon},
  \bibinfo{person}{R Page}, \bibinfo{person}{J Quiroga},
  \bibinfo{person}{Arshad~A Taseen}, {and} \bibinfo{person}{F Tomasini}.}
  \bibinfo{year}{1986}\natexlab{}.
\newblock \showarticletitle{Evaluation of forecasting methods for decision
  support}.
\newblock \bibinfo{journal}{\emph{International journal of forecasting}}
  \bibinfo{volume}{2}, \bibinfo{number}{2} (\bibinfo{year}{1986}),
  \bibinfo{pages}{139--152}.
\newblock


\bibitem[Wu et~al\mbox{.}(2023)]%
        {DBLP:conf/iclr/WuHLZ0L23}
\bibfield{author}{\bibinfo{person}{Haixu Wu}, \bibinfo{person}{Tengge Hu},
  \bibinfo{person}{Yong Liu}, \bibinfo{person}{Hang Zhou},
  \bibinfo{person}{Jianmin Wang}, {and} \bibinfo{person}{Mingsheng Long}.}
  \bibinfo{year}{2023}\natexlab{}.
\newblock \showarticletitle{TimesNet: Temporal 2D-Variation Modeling for
  General Time Series Analysis}. In \bibinfo{booktitle}{\emph{{ICLR}}}.
  \bibinfo{publisher}{OpenReview.net}.
\newblock


\bibitem[Wu et~al\mbox{.}(2021)]%
        {DBLP:conf/nips/WuXWL21}
\bibfield{author}{\bibinfo{person}{Haixu Wu}, \bibinfo{person}{Jiehui Xu},
  \bibinfo{person}{Jianmin Wang}, {and} \bibinfo{person}{Mingsheng Long}.}
  \bibinfo{year}{2021}\natexlab{}.
\newblock \showarticletitle{Autoformer: Decomposition Transformers with
  Auto-Correlation for Long-Term Series Forecasting}. In
  \bibinfo{booktitle}{\emph{NeurIPS}}. \bibinfo{pages}{22419--22430}.
\newblock


\bibitem[Yan and Ouyang(2018)]%
        {DBLP:journals/wpc/YanO18}
\bibfield{author}{\bibinfo{person}{Hongju Yan} {and} \bibinfo{person}{Hongbing
  Ouyang}.} \bibinfo{year}{2018}\natexlab{}.
\newblock \showarticletitle{Financial Time Series Prediction Based on Deep
  Learning}.
\newblock \bibinfo{journal}{\emph{Wirel. Pers. Commun.}} \bibinfo{volume}{102},
  \bibinfo{number}{2} (\bibinfo{year}{2018}), \bibinfo{pages}{683--700}.
\newblock


\bibitem[Zaremba et~al\mbox{.}(2014)]%
        {DBLP:journals/corr/ZarembaSV14}
\bibfield{author}{\bibinfo{person}{Wojciech Zaremba}, \bibinfo{person}{Ilya
  Sutskever}, {and} \bibinfo{person}{Oriol Vinyals}.}
  \bibinfo{year}{2014}\natexlab{}.
\newblock \showarticletitle{Recurrent Neural Network Regularization}.
\newblock \bibinfo{journal}{\emph{CoRR}}  \bibinfo{volume}{abs/1409.2329}
  (\bibinfo{year}{2014}).
\newblock


\bibitem[Zeng et~al\mbox{.}(2023)]%
        {DBLP:conf/aaai/ZengCZ023}
\bibfield{author}{\bibinfo{person}{Ailing Zeng}, \bibinfo{person}{Muxi Chen},
  \bibinfo{person}{Lei Zhang}, {and} \bibinfo{person}{Qiang Xu}.}
  \bibinfo{year}{2023}\natexlab{}.
\newblock \showarticletitle{Are Transformers Effective for Time Series
  Forecasting?}. In \bibinfo{booktitle}{\emph{{AAAI}}}.
  \bibinfo{publisher}{{AAAI} Press}, \bibinfo{pages}{11121--11128}.
\newblock


\bibitem[Zhang and Yan(2023)]%
        {DBLP:conf/iclr/ZhangY23}
\bibfield{author}{\bibinfo{person}{Yunhao Zhang} {and} \bibinfo{person}{Junchi
  Yan}.} \bibinfo{year}{2023}\natexlab{}.
\newblock \showarticletitle{Crossformer: Transformer Utilizing Cross-Dimension
  Dependency for Multivariate Time Series Forecasting}. In
  \bibinfo{booktitle}{\emph{{ICLR}}}. \bibinfo{publisher}{OpenReview.net}.
\newblock


\bibitem[Zhou et~al\mbox{.}(2022)]%
        {DBLP:conf/icml/ZhouMWW0022}
\bibfield{author}{\bibinfo{person}{Tian Zhou}, \bibinfo{person}{Ziqing Ma},
  \bibinfo{person}{Qingsong Wen}, \bibinfo{person}{Xue Wang},
  \bibinfo{person}{Liang Sun}, {and} \bibinfo{person}{Rong Jin}.}
  \bibinfo{year}{2022}\natexlab{}.
\newblock \showarticletitle{FEDformer: Frequency Enhanced Decomposed
  Transformer for Long-term Series Forecasting}. In
  \bibinfo{booktitle}{\emph{{ICML}}} \emph{(\bibinfo{series}{Proceedings of
  Machine Learning Research}, Vol.~\bibinfo{volume}{162})}.
  \bibinfo{publisher}{{PMLR}}, \bibinfo{pages}{27268--27286}.
\newblock


\bibitem[Zhu et~al\mbox{.}(2024)]%
        {DBLP:journals/corr/abs-2401-09417}
\bibfield{author}{\bibinfo{person}{Lianghui Zhu}, \bibinfo{person}{Bencheng
  Liao}, \bibinfo{person}{Qian Zhang}, \bibinfo{person}{Xinlong Wang},
  \bibinfo{person}{Wenyu Liu}, {and} \bibinfo{person}{Xinggang Wang}.}
  \bibinfo{year}{2024}\natexlab{}.
\newblock \showarticletitle{Vision Mamba: Efficient Visual Representation
  Learning with Bidirectional State Space Model}.
\newblock \bibinfo{journal}{\emph{CoRR}}  \bibinfo{volume}{abs/2401.09417}
  (\bibinfo{year}{2024}).
\newblock


\end{thebibliography}
